\pdfoutput=1

\documentclass[11pt]{article}

\usepackage[preprint]{acl}

\usepackage{times}
\usepackage{latexsym}

\usepackage[T1]{fontenc}

\usepackage[utf8]{inputenc}

\usepackage{microtype}

\usepackage{inconsolata}

\usepackage{graphicx}
\usepackage{graphicx}
\usepackage{booktabs}
\usepackage{colortbl}

\usepackage[frozencache, cachedir=.]{minted}
\usepackage{inconsolata}             
\setminted{
  fontsize=\footnotesize,            
  breaklines,
  autogobble,
  tabsize=2,
  numbersep=6pt                      
}

\usepackage{soul}
\usepackage{ulem}
\usepackage{cancel}

\usepackage{amsthm} 
\usepackage{enumitem}
\usepackage{subcaption}
\usepackage{bbm}
\usepackage{amsmath,amssymb,amsfonts}
\usepackage{bm}

\usepackage{eqparbox}
\usepackage{graphicx}
\usepackage{xspace}
\usepackage{comment}
\usepackage{multirow}
\usepackage{verbatim}
\usepackage{arydshln}
\usepackage{tabularx}
\usepackage{tabulary}
\usepackage{booktabs}
\usepackage{makecell}
\usepackage{xspace}
\usepackage{circledsteps}
\usepackage{textcomp}
\usepackage{wasysym}

\usepackage{wrapfig}
\usepackage{algorithm}
\usepackage{algorithmic}

\definecolor{Skyblue}{rgb}{0.6, 0.6, 0.95 }
\definecolor{Green}{rgb}{0.0, 0.8, 0.0 }

\usepackage[hang,flushmargin]{footmisc}
\theoremstyle{definition}

\newcommand{\cmark}{$\bigcirc$}%
\newcommand{\xmark}{\ding{53}}%

\newcommand{\cmmnt}[1]{\ignorespaces} 

\newcommand{\argmax}{\operatornamewithlimits{argmax}}

\newcommand{\algname}{\textsc{WildAgtEval}}
\newcommand{\algurl}{https://github.com/Demon-JieHao/WildAGTEval}

\makeatletter
\newcommand*\bigcdot{\mathpalette\bigcdot@{.5}}
\newcommand*\bigcdot@[2]{\mathbin{\vcenter{\hbox{\scalebox{#2}{$\m@th#1\bullet$}}}}}
\makeatother

\usepackage{hyperref}

\usepackage{amsmath}
\usepackage{amssymb}
\usepackage{mathtools}
\usepackage{amsthm}

\usepackage{balance}
\usepackage{stfloats}

\usepackage{pifont}
\usepackage{xcolor}
\newcommand{\yes}{\textcolor{red}{\textbf{$\bigcirc$}}}
\newcommand{\no}{\textcolor{blue}{\ding{53}}}

\usepackage[most]{tcolorbox}    	
\tcbuselibrary{theorems, skins, }
\newcounter{promptcounter}
\renewcommand{\thepromptcounter}{\arabic{promptcounter}}

\newtcolorbox{MyBox}[2][]{%
  enhanced,
  breakable,
  colback=gray!5,
  colframe=gray!80!black,
  boxrule=1pt,
  toprule=2pt,
  rounded corners,
  arc=2pt,
  top=1.7mm,
  bottom=1.7mm,
  left=2mm,
  right=2mm,
  fuzzy shadow={0pt}{-2pt}{-0.5pt}{0.5pt}{black!35},
  title={\small Example~\thepromptcounter.~#2}, 
  #1 
}

\newtcolorbox{MyBoxWide}[2][]{%
  enhanced,
  breakable,
  colback=gray!5,
  colframe=gray!80!black,
  boxrule=1pt,
  toprule=2pt,
  rounded corners,
  arc=2pt,
  top=1.7mm,
  bottom=1.7mm,
  left=2mm,
  right=2mm,
  fuzzy shadow={0pt}{-2pt}{-0.5pt}{0.5pt}{black!35},
  title={\small Example~\thepromptcounter.~#2},
  float*={t},        
  width=\textwidth,     
  #1
}
\theoremstyle{plain}

\theoremstyle{remark}

\title{Beyond Perfect APIs: A Comprehensive Evaluation of LLM Agents\\Under Real-World API Complexity} 

\author{%
  \textbf{\shortstack[c]{Doyoung Kim\textsuperscript{1} \textsuperscript{2} \quad\quad Zhiwei Ren\textsuperscript{1} \textsuperscript{3} \quad\quad Jie Hao\textsuperscript{1} \textsuperscript{*} \quad\quad Zhongkai Sun\textsuperscript{1} \textsuperscript{*}\quad\quad \\ Lichao Wang\textsuperscript{1} \quad\quad Xiyao Ma\textsuperscript{1} \quad\quad Zack Ye\textsuperscript{1} \quad\quad Xu Han\textsuperscript{1} \quad\quad Jun Yin\textsuperscript{1} \quad\quad Heng Ji\textsuperscript{4} \quad\quad \\ Wei Shen\textsuperscript{1} \quad\quad Xing Fan\textsuperscript{1} \quad\quad Benjamin Yao\textsuperscript{1} \quad\quad Chenlei Guo\textsuperscript{1} \quad\quad }}
  \\\\ \textsuperscript{1}{Amazon} \quad \textsuperscript{2}{KAIST} \quad \textsuperscript{3}{University of Pittsburgh}  \quad \textsuperscript{4}{University of Illinois Urbana-Champaign}
  }

\begin{document}
\maketitle
\begingroup
  \renewcommand\thefootnote{} 
  \footnotetext{\textsuperscript{*}Corresponding authors for this work.}
\endgroup
\setcounter{footnote}{0} 

\begin{abstract}
We introduce \textbf{\algname{}}\footnote{Code \& Data: \href{https://github.com/Demon-JieHao/WildAGTEval}{\texttt{github.com/Demon-JieHao/WildAGTEval}}.}, a benchmark designed to evaluate large language model\,(LLM) agents’ function-calling capabilities under \textit{realistic API complexity}. Unlike prior work that assumes an \textit{idealized} API system and disregards real-world factors such as noisy API outputs, \textbf{\algname{}} accounts for two dimensions of real-world complexity: \textbf{\ding{182}\,API specification}, which includes detailed documentation and usage constraints, and \textbf{\ding{183}\,API execution}, which captures runtime challenges. Consequently, \textbf{\algname{}} offers (i) an API system encompassing 60 distinct complexity scenarios that can be composed into approximately 32K test configurations, and (ii) user-agent interactions for evaluating LLM agents on these scenarios. Using \textbf{\algname{}}, we systematically assess several advanced LLMs and observe that most scenarios are challenging, with \textit{irrelevant information} complexity posing the greatest difficulty and reducing the performance of strong LLMs by 27.3\%. Furthermore, our qualitative analysis reveals that LLMs occasionally \textit{distort} user intent merely to claim task completion, critically affecting user satisfaction.

\end{abstract}

\section{Introduction}
\label{sec:intro}
\definecolor{darkgreen}{RGB}{0,100,0}

\begin{figure}[t]
  \centering
  \begin{subfigure}[t]{\linewidth}
    \centering
    \includegraphics[width=.99\linewidth]{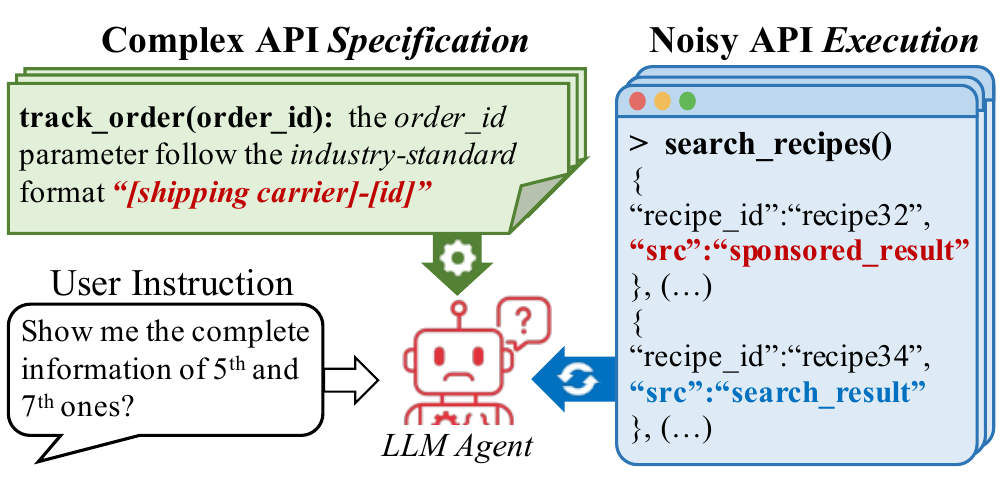}\\[-0.6em]
    \caption{Challenges in real-world agent deployment.}
    \vspace{0.5em}
    \label{fig:exp_a}
  \end{subfigure}
  \begin{subfigure}[t]{\linewidth}
    \centering
    \includegraphics[width=.99\linewidth]{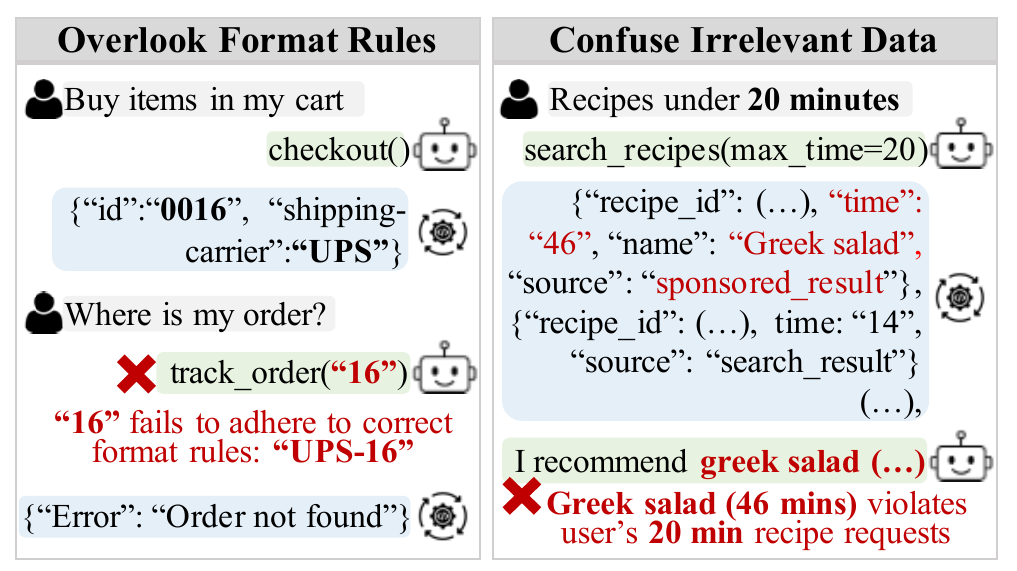}\\[-0.7em]
    \caption{Unassessable agent failures by current benchmarks.}
    \label{fig:exp_b}
  \end{subfigure}
  \vspace{-0.9em} 
  \caption{\textbf{Key motivation for \algname{}:} (a) highlights the challenges in real-world agent deployment; and (b) provides conversations of \algname{} that reveal LLM agents’ failure modes often overlooked by current benchmarks.}
  \vspace{-1.4em} 
  \label{fig:motivation}
\end{figure}

Large language models\,(LLMs) agents, such as Amazon Alexa, have rapidly emerged as powerful interfaces for numerous real-world applications, building on LLMs’ remarkable performance—surpassing human accuracy in college-level mathematics and excelling in high-stakes domains\,\cite{llm_power1,llm_power2,llm_power3}. To evaluate this role, a growing body of work has introduced \textit{function-} or \textit{tool-calling} benchmarks\,\cite{bfcl, tau, zhong2025complexfuncbench, basu2024nestful}, which assess whether agents produce correct API calls that fulfill user instructions. These benchmarks steadily refine our understanding of whether LLM agents can effectively address diverse instructions and execute complex, multi-step tasks.

Despite these efforts, most existing benchmarks assume an \textit{idealized} scenario in which the API functions are straightforward to use, and always produce reliable outputs. However, as shown in Figure~\ref{fig:motivation}(a), \textbf{\textit{these assumptions deviate substantially from real-world scenarios}}. In practical deployments\,{(e.g., Amazon Alexa)}, agents must carefully adhere to extensive, \textbf{meticulous API specifications}\,(e.g., \textit{domain-specific} formatting rules ``\texttt{[shipping~carrier]-[id]}'') while also managing \textbf{imperfect API execution}, which often produces \textit{noisy} outputs\,(e.g., ``\texttt{sponsored\_result}'') or encounters runtime \textit{errors}.

\begin{table*}[t!]
\centering
\renewcommand{\arraystretch}{0.85}
\setlength{\tabcolsep}{6pt}
\resizebox{0.9999\textwidth}{!}{
\begin{tabular}{@{}l|cccc|cccc@{}}
\toprule
\multirow{4}{*}{\textbf{Benchmarks}} 
& \multicolumn{4}{c|}{\textbf{\ding{182} \textsc{API Specification Complexity}}} 
& \multicolumn{4}{c}{\textbf{\ding{183} \textsc{API Execution Complexity}}} \\
\cmidrule(lr){2-5} \cmidrule(lr){6-9}
& \textit{\begin{tabular}[c]{@{}c@{}}Ad-hoc\\Rules\end{tabular}} 
& \textit{\begin{tabular}[c]{@{}c@{}}Unclear\\Functionality\\Boundaries\end{tabular}} 
& \textit{\begin{tabular}[c]{@{}c@{}}Functional\\Dependencies\end{tabular}} 
& \textit{\begin{tabular}[c]{@{}c@{}}Ambiguous\\Descriptions\end{tabular}} 
& \textit{\begin{tabular}[c]{@{}c@{}}Informational\\Notices\end{tabular}} 
& \textit{\begin{tabular}[c]{@{}c@{}}Partially\\Irrelevant\\Information\end{tabular}} 
& \textit{\begin{tabular}[c]{@{}c@{}}Feature\\Limitation\\Errors\end{tabular}} 
& \textit{\begin{tabular}[c]{@{}c@{}}System\\Failure\\Errors\end{tabular}} \\
\midrule
BFCLv3\,\citep{bfcl} & \cellcolor{red!10}\xmark & \cellcolor{red!10}\xmark & \cellcolor{blue!10}\cmark & \cellcolor{blue!10}\cmark & \cellcolor{red!10}\xmark & \cellcolor{red!10}\xmark & \cellcolor{red!10}\xmark & \cellcolor{blue!10}\cmark \\
ComplexFuncBench\,\cite{zhong2025complexfuncbench} & \cellcolor{red!10}\xmark & \cellcolor{red!10}\xmark & \cellcolor{blue!10}\cmark & \cellcolor{blue!10}\cmark & \cellcolor{red!10}\xmark & \cellcolor{red!10}\xmark & \cellcolor{red!10}\xmark & \cellcolor{red!10}\xmark \\
$\tau$-bench\,\cite{tau} & \cellcolor{red!10}\xmark & \cellcolor{red!10}\xmark & \cellcolor{blue!10}\cmark & \cellcolor{red!10}\xmark & \cellcolor{red!10}\xmark & \cellcolor{red!10}\xmark & \cellcolor{red!10}\xmark & \cellcolor{red!10}\xmark \\
ToolSandbox\,\cite{tau} & \cellcolor{red!10}\xmark & \cellcolor{red!10}\xmark & \cellcolor{blue!10}\cmark & \cellcolor{red!10}\xmark & \cellcolor{red!10}\xmark & \cellcolor{red!10}\xmark & \cellcolor{red!10}\xmark & \cellcolor{blue!10}\cmark \\
AgentGym\,\cite{agentgym} & \cellcolor{red!10}\xmark & \cellcolor{red!10}\xmark & \cellcolor{blue!10}\cmark & \cellcolor{red!10}\xmark & \cellcolor{red!10}\xmark & \cellcolor{red!10}\xmark & \cellcolor{red!10}\xmark & \cellcolor{red!10}\xmark \\
Multi-Turn-Instruct\,\cite{multi-turn-inst} & \cellcolor{red!10}\xmark & \cellcolor{red!10}\xmark & \cellcolor{red!10}\xmark & \cellcolor{red!10}\xmark & \cellcolor{red!10}\xmark & \cellcolor{red!10}\xmark & \cellcolor{red!10}\xmark & \cellcolor{red!10}\xmark \\
Lost-in-Conv\,\cite{lostinconv} & \cellcolor{red!10}\xmark & \cellcolor{red!10}\xmark & \cellcolor{red!10}\xmark & \cellcolor{red!10}\xmark & \cellcolor{red!10}\xmark & \cellcolor{red!10}\xmark & \cellcolor{red!10}\xmark & \cellcolor{red!10}\xmark \\
\midrule
\textbf{\algname{}} & \cellcolor{blue!10}\cmark & \cellcolor{blue!10}\cmark & \cellcolor{blue!10}\cmark & \cellcolor{blue!10}\cmark & \cellcolor{blue!10}\cmark & \cellcolor{blue!10}\cmark & \cellcolor{blue!10}\cmark & \cellcolor{blue!10}\cmark \\
\bottomrule
\end{tabular}
}
\vspace{-0.2cm}
\caption{Comparison of \textit{API complexity coverage} between prior representative benchmarks and \algname{}.}
\label{tab:contribution}
\vspace{-0.4cm}
\end{table*}

Consequently, current benchmarks often produce \textit{overly} optimistic capability assessments by failing to evaluate agent performance under realistic complexities. For example, in Figure~\ref{fig:motivation}(b), these benchmarks cannot detect agent failures arising from intricate \textit{API specification}, wherein agents simply use seemingly relevant information\,(e.g., ``16'') rather than adhering to the required format\,(e.g., ``UPS-16''). Similarly, they do not capture failures stemming from noisy \textit{API execution results}, such as when agents recommend inappropriate content\,(e.g., a 46-minute recipe for a 20-minute meal request) due to confusion over irrelevant sponsored results.

To address this gap, we propose \textbf{\algname{}}, a novel benchmark that moves beyond idealized APIs to evaluate LLM agents’ ability to invoke external functions under \textit{\textbf{real-world API complexity}}. Specifically, \algname{} simulates these real-world complexities within an API system---a fixed suite of functions---thus exposing challenges during user–agent conversations\,(Figure~\ref{fig:motivation}(b)). Consequently, \algname{} provides (i) the API system and (ii) user-agent interactions grounded in it, covering a broad range of \textit{complexity types}, as shown in Table~\ref{tab:contribution}: \textbf{\ding{182}\,\textit{API specification}}, covering intricate documentation and usage rules, and \textbf{\ding{183}\,\textit{API execution}}, capturing runtime challenges. Across these dimensions, the API system includes 60 specific \textit{complexity scenarios}, yielding approximately 32K distinct test configurations. User-agent interactions for these scenarios are generated using a recent conversation-generation method\,\cite{barres2025tau}.

An important consideration is that the complexity scenarios in the API system should faithfully reflect the real-world API environments. To this end, we employ a novel \textit{assign-and-inject} mechanism that integrates complexities into the API system, leveraging the insight that each complexity type naturally arises in specific categories of API functions based on their functionalities. For example, {\textit{irrelevant information}} frequently occurs in information-retrieval functions\,(e.g., \texttt{search\_recipes()} in Figure~\ref{fig:motivation}(b)). Accordingly, we first \textit{assign} each complexity type to the functions most likely to encounter this type of complexity in the real world; and then \textit{inject} these complexities by modifying the corresponding API implementations.

Our evaluation on \algname{} shows that most complexity scenarios consistently degrade performance across strong LLM agents\,(e.g., Claude-4-Sonnet\,\cite{Claude4_Sonnet}), with \textit{irrelevant information} complexity posing the greatest challenge, causing an average performance drop of 27.3\%. Moreover, when multiple complexities accumulate, the performance degrades by up to 63.2\%. Qualitative analysis further reveals that, when facing unresolvable tasks, LLMs persist in attempting to solve them, ultimately distorting user intent and producing misleading success responses.


\section{Related Work}
\label{sec:related_works}

\subsection{API-Based Benchmarks for LLM Agents}
\label{subsec:API IFs}

Existing API-based benchmarks have advanced agent evaluation by focusing on multi-step reasoning through functional dependencies. BFCLv3\,\cite{bfcl} and $\tau$-bench\,\cite{tau, barres2025tau} introduce sequential dependency scenarios\,(e.g., \texttt{search\_media()} 
→ \texttt{play()} for playback). Moreover, ComplexFuncBench\,\cite{zhong2025complexfuncbench} incorporates user constraints into multi-step tasks and NESTful\,\cite{basu2024nestful} extends to mathematical domains. Meanwhile, Incomplete-APIBank\,\cite{syserror1} and ToolSandbox\,\cite{toolsandbox} assess robustness of agents to missing APIs and service state variations, respectively. Nevertheless, prior benchmarks still underrepresent real-world API complexities\,(Table~\ref{tab:contribution}), yielding overly optimistic assessments; this work addresses this gap by integrating such complexities into agent evaluation.

\subsection{User-Based Benchmarks for LLM Agents}
\label{subsec:user IFs}
User-based benchmarks assess whether LLM agents fulfill diverse real-world user instructions of varying complexity. MT-Eval\,\cite{mt-eval}, Multi-Turn-Instruct\,\cite{multi-turn-inst}, and Lost-In-Conv\,\cite{lostinconv} evaluate how well agents handle diverse user instructions in multi-turn interactions, emphasizing challenges such as intent disambiguation, multi-intent planning, and preference elicitation. IHEval\,\cite{zhang2025iheval} and UserBench\,\cite{userbench} further extend these evaluations by incorporating scenarios where user intents are conflicting and evolving, respectively. Despite their broad coverage, they typically overlook the complexities related to tool invocation, such as complex API specifications or noisy API execution.

\def\arraystretch{0.8}
\renewcommand\cellalign{tl} 

\definecolor{specgray}{gray}{0.92}
\definecolor{execgray}{gray}{0.96}

\definecolor{spec}{HTML}{009E73}
\definecolor{specbg}{HTML}{EEF6E9}
\definecolor{exec}{HTML}{0072B2}
\definecolor{execbg}{HTML}{EBF2FA}
\definecolor{sponsored}{HTML}{D55E00}
\definecolor{advice}{HTML}{E69F00}
\definecolor{searchc}{HTML}{56B4E9}
\definecolor{ambig}{HTML}{CC79A7}
\newcolumntype{Y}{>{\centering\arraybackslash}m{1.9cm}}

\newcommand{\Prompt}{\textcolor{spec}{\textsc{}}}
\newcommand{\Result}{\textcolor{exec}{\textsc{}}}
\newcommand{\PromptCap}{\textcolor{spec}{\textsc{System Prompt}}}
\newcommand{\ResultCap}{\textcolor{exec}{\textsc{Execution Result}}}
\newcommand{\ex}[1]{\footnotesize #1}
\newcommand{\exspec}[1]{\cellcolor{specbg}\Prompt\ \ex{#1}}
\newcommand{\exexec}[1]{\cellcolor{execbg}\Result\ \ex{#1}}

\begin{table*}[t!]
\centering
\footnotesize
\setlength{\tabcolsep}{5pt}
\resizebox{0.9999\linewidth}{!}{%
\begin{tabularx}{\textwidth}{@{}c p{1.8cm}| X X p{0.25\linewidth} p{0.1\linewidth}@{}}
\arrayrulecolor{black}\specialrule{1.3pt}{1.0pt}{1.0pt}
\textbf{} &
\thead{\textbf{Complexity}\\\textbf{Type}} &
\thead{\textbf{Description}} &
\thead{\textbf{Occurrence}\\\textbf{Context}} &
\thead{\textbf{Example}} &
\thead{\textbf{Desirable}\\\textbf{Action}}\\
\midrule

\multirow[c]{4}{*}{\vspace{-10em}\rotatebox{90}{\textsc{API Specification}}} &
{\cellcolor{specbg} \textbf{Ad-hoc rules}} &
Enforces domain- or legacy-driven formats or usage conventions &
Imposed by domain-specific standards, legacy design, or regulation &
{Parameter ``time'' requires ISO 8601 format\,\cite{iso8601}; parameter ``phone\_number'' requires E.164 format\,\cite{ituE164}} &
Follow required format\\
\addlinespace[0.5ex]\cdashline{2-6}\addlinespace[0.7ex]

& {\cellcolor{specbg} \textbf{Unclear functionality boundaries}} &
Exposes superficially similar functions with distinct functionalities &
Caused by incremental integration of API functions without a coordinated naming convention &
{Includes similarly named API functions with different roles: \texttt{search\_product()} and \texttt{search\_inventory()}} &
Predict correctly without confusion\\
\addlinespace[0.5ex]\cdashline{2-6}\addlinespace[0.7ex]

& {\cellcolor{specbg} \textbf{Functional dependencies}} &
Requires a fixed sequence of prerequisite API calls &
Arises from underlying workflow of API system &
{Devices must be powered on before being used} &
Orchestrate dependency\\
\addlinespace[0.5ex]\cdashline{2-6}\addlinespace[0.7ex]

& {\cellcolor{specbg} \textbf{Ambiguous descriptions}} &
Omits specification of units, defaults, etc. &
Caused by inconsistent documentation updates  &
{Parameter ``temperature'' omits the unit\,(e.g., °C or °F)} &
Infer sensible default\\
\addlinespace[0.5ex]\cline{1-6}\addlinespace[0.7ex]

\multirow{4}{*}{\vspace{-11.5em}\rotatebox[origin=c]{90}{\textsc{API Execution}}} &
{\cellcolor{execbg} \textbf{Informational notices}} &
Returns successful API results with  advisory messages\,(warnings, or companion actions) &
Provided to promote new features or deliver best-practice guidance &
{Displays an informational notice: ``Frequently used operations: \texttt{brightness\_adjust()},  \texttt{color\_set()},  \texttt{play()}, etc.''} &
Proceed without being affected by notices\\
\addlinespace[0.5ex]\cdashline{2-6}\addlinespace[0.7ex]

& {\cellcolor{execbg} \textbf{Partially irrelevant information}} &
Returns requested data with irrelevant content such as advertisements &
Introduced by monetization or non-essential tracking information &
{Insert sponsored companies’ contents into the main results} &
Filter out irrelevant information\\
\addlinespace[0.5ex]\cdashline{2-6}\addlinespace[0.7ex]

& {\cellcolor{execbg} \makecell{\textbf{Feature}\\\textbf{limitation}\\\textbf{errors}} } &
Fails requests due to feature unavailability accompanied by workarounds &
Triggered by limited quotas, size caps, or plan entitlements &
{API request fails and displays implicit workaround: ``search is only limited to recent info.''} &
{Apply work-around (`recent search')}\\
\addlinespace[0.5ex]\cdashline{2-6}\addlinespace[0.7ex]

& {\cellcolor{execbg} \makecell{\textbf{System}\\\textbf{failure errors}} } &
Fails completely due to system issues &
Caused by infrastructure or upstream outages &
{API request fails and displays system-side cryptic error codes} &
Provide error reporting\\
\arrayrulecolor{black}\specialrule{1.3pt}{1.0pt}{1.0pt}
\end{tabularx}
}
\vspace{-0.3cm}
\caption{API complexity taxonomy. Each entry lists its occurrence context, a representative example, and the desirable agent action. The representative agent failure cases are discussed in Section~\ref{subsec:core_failures}.}
\vspace{-0.5cm}
\label{tab:api_complexity_taxonomy}
\end{table*}

\section{\algname{}: Evaluating LLM Agents under API Complexity}
\label{sec:benchmark}
We present \algname{} to benchmark the robustness of LLM agents when invoking external functions under \textit{real-world API complexities}. Following prior work\,\cite{bfcl, zhong2025complexfuncbench}, an agent receives (i) an executable API system and (ii) a sequence of user-agent interactions\,(hereafter, referred to as the
\textit{conversations}), and then produces responses by invoking the appropriate API calls.

\algname{} advances existing benchmarks to better reflect real-world agent challenges. It introduces eight \textbf{API complexity types} commonly observed in practice, and \textbf{integrates them into an API system} guided by real-world usage patterns. As a result, \algname{} comprises 60 distinct complexity scenarios, supporting a potential of approximately 32,000 unique test configurations.

\subsection{Taxonomy of API Complexities}
Table~\ref{tab:api_complexity_taxonomy} describes the eight types of API complexity spanning the API specification and execution phases, and Figure~\ref{fig:prompt} illustrates a prompt for an agent that shows how these complexities manifest in each API function. 
The \textit{specification complexities} affect what the agent reads. Within the agent's prompt, these complexities appear in the ``API functions'' and ``Instructions'' sections\,(see Figure~\ref{fig:prompt}), ensuring their consistent inclusion during prompt construction. In contrast, the \textit{execution complexities} affect what the agent observes after making API calls; they are introduced into subsequent prompts according to the API calls invoked by the agent. As shown by [C6] in Figure~\ref{fig:prompt}, once the agent calls \texttt{stock\_price()}, the subsequent prompt contains irrelevant information\,(e.g., ``AAPL\,(sponsored)''), forcing the agent to filter or reconcile noisy outputs.

\subsection{Real-World Complexity Integration}
Each complexity type\,(e.g., {irrelevant information}) naturally arises in specific categories of functions, reflecting their core functionalities\,(e.g., information retrieval). Building on this observation, we employ an \textit{assign-and-inject} complexity integration mechanism where each complexity type is first \textit{assigned} to the functions most likely to encounter this type of complexity---based on its real-world likehood of occurrence---and subsequently \textit{injected} into those functions. For example, the \textit{irrelvant information} complexity is assigned and injected into the information retrieval function, \texttt{stock\_price()}. Furthermore, to ensure natural assignments of all complexity types to relevant functions, we construct our API system to include a sufficiently diverse set of functions.

\section{Benchmark Construction through Complexity Integration}
\label{sec:construction}

\definecolor{textolive}{RGB}{55,87,34} 
\definecolor{deepblue}{RGB}{40,86,126} 

\begin{figure*}[t!]
    \centering
    \includegraphics[width=\textwidth]{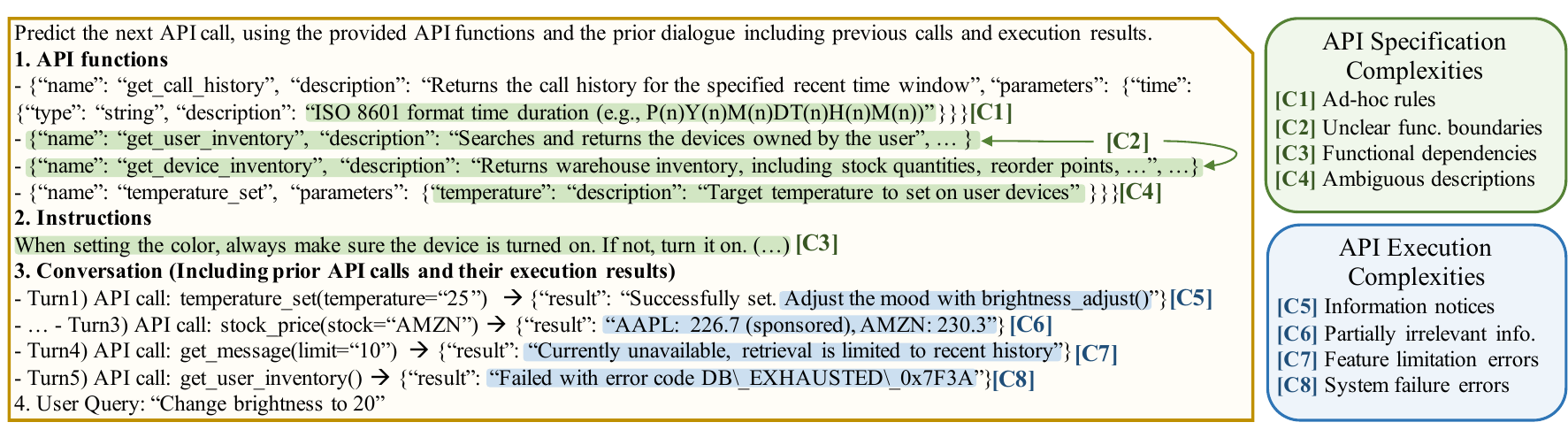}
    \vspace*{-0.9cm}
    \caption{A prompt for an LLM agent in \algname{}, encompassing \textcolor{textolive}{\textit{specification-level complexities}}---conventional parameter rules, similar yet functionally distinct functions, domain-specific dependencies, and undocumented details\,(\textcolor{textolive}{[C1–C4]})---alongside \textcolor{deepblue}{\textit{execution-level complexities}}---companion function notice, irrelevant sponsored content, partial failure with implicit workaround, and complete failure with cryptic error code\,(\textcolor{deepblue}{[C5–C8]}).
    } 
    \label{fig:prompt}
    \vspace*{-0.6cm}
\end{figure*}

\begin{figure}[!t]
    \centering
    \includegraphics[width=1.00\linewidth]{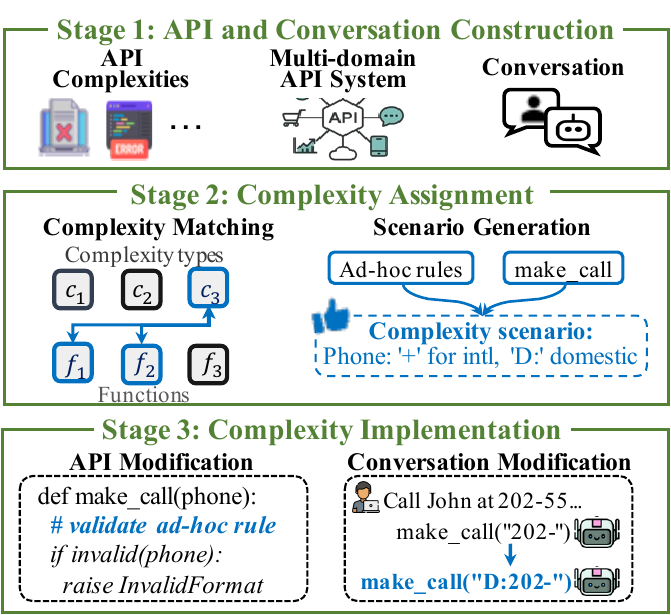}
    \vspace{-0.75cm}
    \caption{Overview of \algname{} construction.}
    \label{fig:overview}
    \vspace*{-0.6cm}
\end{figure}

\subsection{Overview}
\label{subsec:overview}
Figure~\ref{fig:overview} illustrates the \textit{assign-and-inject} process that constructs \textbf{\algname{}}, outlined as 
\begin{itemize}[leftmargin=*, noitemsep, topsep=0pt]
\item \textbf{Stage~1:} Construct a multi-domain API system and conversations grounded in that system.
\item \textbf{Stage~2:} \textit{Assign} complexities to relevant functions and create concrete complexity scenarios.
\item \textbf{Stage~3:} \textit{Inject} these scenarios into API functions and, when required, into conversations.
\end{itemize}

\subsection{Stage 1: Multi-Domain API System and Conversation Construction}
\noindent \textbf{API system construction.} To enable natural complexity integration across all complexity types, we develop a comprehensive multi-domain API system spanning seven commonly used domains\,(e.g., device control, information retrieval) comprising 86 API functions. Each function is fully executable, accompanied by the relevant databases and policies. Due to the extensive domain coverage and policy constraints, the prompt describing API usage for LLM agents reaches approximately 34K tokens. 

\smallskip
\noindent \textbf{Conversation construction.} 
We adopt the conversation generation framework\,\cite{barres2025tau} to produce multi-turn conversations paired with precise API call annotations. As outlined in Figure~\ref{fig:conv_gen}, the process begins by curating \textit{verified intent primitives}, which are atomic units encapsulating a specific user goal\,(e.g., ``watch a movie'') along with the corresponding API calls\,(e.g., \texttt{search\_media()}, \texttt{power\_on()}, \texttt{play()}). Subsequently, an LLM composes these primitives into longer interaction scenarios, ensuring realistic conversations. All synthesized conversations undergo post-generation validation to confirm both the coherence of the conversational context and the correctness of the associated API call annotations. The resulting dataset contains 300 multi-turn conversations requiring a total of 3{,}525 agent API calls, with each conversation averaging approximately 4.7 user-agent turns and 2.5 API calls per turn.


Further details regarding the construction of both the API system and the conversations are provided in Appendix~\ref{subsec:s1_app}.

\subsection{Stage 2: Complexity Assignment}
We identify \textit{relevant} complexity-function pairs by evaluating their \textit{likelihood of real-world occurrence}, subsequently generating concrete scenarios for pairs deemed highly relevant.

\noindent \textbf{Relevance-based complexity matching.}
Consider a multi-domain API system with functions $\mathcal{F} = \{f_1, ..., f_n\}$ and a set of complexity types $\mathcal{C}$. For each complexity type $c \in \mathcal{C}$, we aim to find a subset of functions $\mathcal{F}^*_c \subset \mathcal{F}$ most relevant to $c$\,(e.g., $\mathcal{F}^*_{c_3}=\{f_1,f_2\}$ in Figure~\ref{fig:overview}).
Specifically, for every pair $(c, f)$, we quantify \textit{likelihood of real-world occurrence} via a \textit{relevance scores} ${\rm r}_{c,f}$. We obtain this score by using an instruction-following language model with a relevance assessment template $\mathcal{I}_{\text{rel}}$ as
\begin{equation}
{\rm r}_{c,f} = {\rm InstructLM}(c, f, \mathcal{I}_{\text{rel}}),
\label{eq:rel}
\end{equation}
where $\mathcal{I}_{\text{rel}}$ specifies how to assess the likelihood that $f$ will exhibit complexity type $c$ in real-world environments\,(see Figure~\ref{fig:prompt_relevance}).

Then, to select the most relevant functions $\mathcal{F}^*_c$ for complexity type $c$, we take the top-$k$ functions based on their relevance score as
\begin{equation}
\mathcal{F}^*_c = \{f : f \in {\rm top}\text{-}{\rm k}({\rm r}_{c,f}) \}.
\end{equation}

\noindent \textbf{Complexity scenario generation.}
After determining the most relevant complexity-function pairs, we generate concrete complexity scenarios (e.g., an ad-hoc rule on the phone number in Figure~\ref{fig:overview}). 
Specifically, for each pair $(c, f)$ with $f \in \mathcal{F}^*_c$, we construct a scenario ${\rm s}_{c,f}$ by mapping the complexity type $c$ to a real-world application context for $f$ using the scenario specification template $\mathcal{I}_{\text{scen}}$ as 
\begin{equation}
\label{eq:scenario_generation}
{\rm s}_{c,f} = {\rm InstructLM}(c, f, \mathcal{I}_{\text{scen}}),
\end{equation}
where $\mathcal{I}_{\text{scen}}$ specifies how $c$ could manifest in $f$ under real-world conditions, as detailed in Figure~\ref{fig:prompt_scenario}. 
Because the generated scenarios exhibit considerable variation, we generate multiple candidates $\{s_{c,f}^{(i)}\}_i$ via Eq.~\eqref{eq:scenario_generation} and select the most representative using a validation template $\mathcal{I}_{\text{val}}$ as
\begin{equation}
{\rm s}_{c,f}^* = \argmax_{{\rm s} \in \{{\rm s}_{c,f}^{(i)}\}_i} {\rm InstructLM}({\rm s}, c, f, \mathcal{I}_{\text{val}}),
\label{eq:val}
\end{equation}
where $\mathcal{I}_{\text{val}}$ evaluates the realism and fidelity to the specified complexity type, as shown in Figure~\ref{fig:prompt_scenario_val}.

\subsection{Stage 3: Complexity Implementation}
We integrate selected scenarios ${\rm s}_{c,f}^*$ into the API function $f$ and into the conversation, adhering to rigorous quality assurance protocols. Table~\ref{tab:complexity_stats} summarizes the complexity scenarios implemented for each complexity type in \algname{}.

\noindent \textbf{API system update.}
We modify the code implementations of relevant API functions to realize the complexity scenarios. For example, we incorporate validation procedures\,(e.g., \texttt{invalid(phone)} in Figure~\ref{fig:overview}) and adjust the function outputs to introduce noise\,(e.g., sponsored content).

\noindent \textbf{Conversation update.}
Certain complex scenarios require modifications to the annotations for conversations, i.e., gold API calls and reference responses, to accurately reflect newly introduced complexities. For example, when applying ad-hoc rules, we update the gold API calls to follow the required formats\,(e.g., ``\texttt{D:}'' in Figure~\ref{fig:overview}). For feature limitation and system failure errors, we annotate the prescribed workaround and clear error messages, respectively, as the \textit{new} {reference} response. Additional details appear in Appendix~\ref{subsec:s3_app}.

\noindent \textbf{Quality assurance.}
Maintaining API executability and accurate labels is essential for reliable evaluation; for example, if alternative solutions exist, agents may exploit unintended solution paths, potentially undermining the evaluation’s validity. To prevent such issues, we employ (1) manual editing of API functions and conversation labels\,\cite{bfcl, prabhakar2025apigen}, and (2) trial runs with Claude-4-Sonnet\,\cite{Claude4_Sonnet} to detect all alternative solutions, following the prior work\,\cite{tau}.

\section{Evaluation}
\label{sec:evaluation}
\newcommand{\Pad}{\rule{0pt}{2.4ex}\rule[-0.6ex]{0pt}{0pt}}

\def\arraystretch{1.05}
\begin{table*}[t!]
\small
\centering
\resizebox{0.99\linewidth}{!}{%
\begin{tabular}[c]
{@{}c|cc|cc|cc|cc|ccc@{}}
\arrayrulecolor{black}\specialrule{1.3pt}{0.75pt}{1.0pt}
\multicolumn{1}{c|}{\multirow{3}{*}{\hspace{-0.0cm}\makecell[c]{\textbf{LLM Agents}}\hspace{-0.0cm}}}
& \multicolumn{4}{c|}{{ \textsc{Specification Complexity}}} 
& \multicolumn{4}{c|}{{ \textsc{Execution Complexity}}}
& \multicolumn{2}{c}{ \multirow{2}{*}{\makecell[c]{%
      \cellcolor{gray!10}\Pad\textsc{Average\,(\%)~~}\\[-0.2ex]
      \cellcolor{gray!10}\Pad(Agent-wise)~~%
    } } }
\\ \addlinespace[0.5ex] 
& \multicolumn{2}{c|}{\textbf{Ad-hoc Rules}}
& \multicolumn{2}{c|}{\hspace{-0.0cm}\textbf{Unclear Func.}\hspace{-0.3cm}}
& \multicolumn{2}{c|}{~~\textbf{Info. Notice}~~}
& \multicolumn{2}{c|}{\hspace{-0.cm}\textbf{Irrelvant Info.}\hspace{-0.3cm}}
\\ \addlinespace[0.3ex] 
& \multicolumn{1}{c}{~\no~}
& \multicolumn{1}{c|}{~\yes~}
& \multicolumn{1}{c}{~\no~}
& \multicolumn{1}{c|}{~\yes~}
& \multicolumn{1}{c}{~\no~} 
& \multicolumn{1}{c|}{~\yes~}
& \multicolumn{1}{c}{~\no~}
& \multicolumn{1}{c|}{~\yes~}
& \multicolumn{1}{c}{\cellcolor{gray!10}~\no~}
& \multicolumn{1}{c}{\cellcolor{gray!10}~\yes~}
\\ 
\arrayrulecolor{black}\specialrule{1pt}{1.5pt}{0.7pt} 
\arrayrulecolor{black}\specialrule{1pt}{0.7pt}{3.0pt}
\multicolumn{1}{c|}{{ \texttt{Claude-4.0-Sonnet\,(Think)} } } 
& \begin{tabular}[c]{@{}c@{}}{{78.3}}\end{tabular}
& \begin{tabular}[c]{@{}c@{}}{{69.6}}\end{tabular}
& \begin{tabular}[c]{@{}c@{}}{{63.3}}\end{tabular}
& \begin{tabular}[c]{@{}c@{}}{{56.9}}\end{tabular}
& \begin{tabular}[c]{@{}c@{}}{{87.9}}\end{tabular}
& \begin{tabular}[c]{@{}c@{}}{{81.8}}\end{tabular}
& \begin{tabular}[c]{@{}c@{}}{{71.2}}\end{tabular}
& \begin{tabular}[c]{@{}c@{}}{{61.6}}\end{tabular}
& \cellcolor{gray!10}\begin{tabular}[c]{@{}c@{}}{{75.2}}\end{tabular}
& \cellcolor{gray!10}\begin{tabular}[c]{@{}c@{}}{{67.5}}\end{tabular}
\\ 
\multicolumn{1}{c|}{{ \texttt{Claude-4.0-Sonnet}  }} 
& \begin{tabular}[c]{@{}c@{}}{{76.1}}\end{tabular}
& \begin{tabular}[c]{@{}c@{}}{{69.6}}\end{tabular}
& \begin{tabular}[c]{@{}c@{}}{{55.0}}\end{tabular}
& \begin{tabular}[c]{@{}c@{}}{{48.6}}\end{tabular}
& \begin{tabular}[c]{@{}c@{}}{{86.4}}\end{tabular}
& \begin{tabular}[c]{@{}c@{}}{{81.8}}\end{tabular}
& \begin{tabular}[c]{@{}c@{}}{{68.8}}\end{tabular}
& \begin{tabular}[c]{@{}c@{}}{{54.5}}\end{tabular}
& \cellcolor{gray!10}\begin{tabular}[c]{@{}c@{}}{{71.6}}\end{tabular}
& \cellcolor{gray!10}\begin{tabular}[c]{@{}c@{}}{{63.6}}\end{tabular}
\\ 
\multicolumn{1}{c|}{{ \texttt{Claude-3.7-Sonnet} }} 
& \begin{tabular}[c]{@{}c@{}}{{78.3}}\end{tabular}
& \begin{tabular}[c]{@{}c@{}}{{54.3}}\end{tabular}
& \begin{tabular}[c]{@{}c@{}}{{59.6}}\end{tabular}
& \begin{tabular}[c]{@{}c@{}}{{56.9}}\end{tabular}
& \begin{tabular}[c]{@{}c@{}}{{81.8}}\end{tabular}
& \begin{tabular}[c]{@{}c@{}}{{81.8}}\end{tabular}
& \begin{tabular}[c]{@{}c@{}}{{67.4}}\end{tabular}
& \begin{tabular}[c]{@{}c@{}}{{53.5}}\end{tabular}
& \cellcolor{gray!10}\begin{tabular}[c]{@{}c@{}}{{71.8}}\end{tabular}
& \cellcolor{gray!10}\begin{tabular}[c]{@{}c@{}}{{61.6}}\end{tabular}
\\
\multicolumn{1}{c|}{{ \texttt{Claude-3.5-Sonnet} }} 
& \begin{tabular}[c]{@{}c@{}}{{67.4}}\end{tabular}
& \begin{tabular}[c]{@{}c@{}}{{60.9}}\end{tabular}
& \begin{tabular}[c]{@{}c@{}}{{40.4}}\end{tabular}
& \begin{tabular}[c]{@{}c@{}}{{40.4}}\end{tabular}
& \begin{tabular}[c]{@{}c@{}}{{75.0}}\end{tabular}
& \begin{tabular}[c]{@{}c@{}}{{70.5}}\end{tabular}
& \begin{tabular}[c]{@{}c@{}}{{62.8}}\end{tabular}
& \begin{tabular}[c]{@{}c@{}}{{51.2}}\end{tabular}
& \cellcolor{gray!10}\begin{tabular}[c]{@{}c@{}}{{61.4}}\end{tabular}
& \cellcolor{gray!10}\begin{tabular}[c]{@{}c@{}}{{55.8}}\end{tabular}
\\ \addlinespace[0.5ex]\cdashline{1-11}\addlinespace[0.7ex]
\multicolumn{1}{c|}{{ \texttt{GPT-OSS-120B} }} 
& \begin{tabular}[c]{@{}c@{}}{{73.9}}\end{tabular}
& \begin{tabular}[c]{@{}c@{}}{{65.2}}\end{tabular}
& \begin{tabular}[c]{@{}c@{}}{{52.3}}\end{tabular}
& \begin{tabular}[c]{@{}c@{}}{{49.5}}\end{tabular}
& \begin{tabular}[c]{@{}c@{}}{{81.8}}\end{tabular}
& \begin{tabular}[c]{@{}c@{}}{{81.8}}\end{tabular}
& \begin{tabular}[c]{@{}c@{}}{{65.1}}\end{tabular}
& \begin{tabular}[c]{@{}c@{}}{{53.5}}\end{tabular}
& \cellcolor{gray!10}\begin{tabular}[c]{@{}c@{}}{{68.3}}\end{tabular}
& \cellcolor{gray!10}\begin{tabular}[c]{@{}c@{}}{{62.5}}\end{tabular}
\\ \addlinespace[0.5ex]\cdashline{1-11}\addlinespace[0.7ex]
\multicolumn{1}{c|}{{ \texttt{Qwen3-235B-Instruct} }} 
& \begin{tabular}[c]{@{}c@{}}{{71.7}}\end{tabular}
& \begin{tabular}[c]{@{}c@{}}{{69.6}}\end{tabular}
& \begin{tabular}[c]{@{}c@{}}{{56.9}}\end{tabular}
& \begin{tabular}[c]{@{}c@{}}{{50.5}}\end{tabular}
& \begin{tabular}[c]{@{}c@{}}{{79.5}}\end{tabular}
& \begin{tabular}[c]{@{}c@{}}{{77.3}}\end{tabular}
& \begin{tabular}[c]{@{}c@{}}{{58.1}}\end{tabular}
& \begin{tabular}[c]{@{}c@{}}{{37.2}}\end{tabular}
& \cellcolor{gray!10}\begin{tabular}[c]{@{}c@{}}{{66.6}}\end{tabular}
& \cellcolor{gray!10}\begin{tabular}[c]{@{}c@{}}{{58.7}}\end{tabular}
\\ 
\multicolumn{1}{c|}{{ \texttt{Qwen3-235B-Thinking} }} 
& \begin{tabular}[c]{@{}c@{}}{{71.7}}\end{tabular}
& \begin{tabular}[c]{@{}c@{}}{{65.2}}\end{tabular}
& \begin{tabular}[c]{@{}c@{}}{{63.3}}\end{tabular}
& \begin{tabular}[c]{@{}c@{}}{{56.9}}\end{tabular}
& \begin{tabular}[c]{@{}c@{}}{{86.4}}\end{tabular}
& \begin{tabular}[c]{@{}c@{}}{{81.8}}\end{tabular}
& \begin{tabular}[c]{@{}c@{}}{{65.1}}\end{tabular}
& \begin{tabular}[c]{@{}c@{}}{{46.5}}\end{tabular}
& \cellcolor{gray!10}\begin{tabular}[c]{@{}c@{}}{{71.6}}\end{tabular}
& \cellcolor{gray!10}\begin{tabular}[c]{@{}c@{}}{{62.6}}\end{tabular}
\\ \addlinespace[0.5ex]\cdashline{1-11}\addlinespace[0.7ex]
\multicolumn{1}{c|}{{ \texttt{Qwen3-32B}}} 
& \begin{tabular}[c]{@{}c@{}}{{56.5}}\end{tabular}
& \begin{tabular}[c]{@{}c@{}}{{54.3}}\end{tabular}
& \begin{tabular}[c]{@{}c@{}}{{36.5}}\end{tabular}
& \begin{tabular}[c]{@{}c@{}}{{35.3}}\end{tabular}
& \begin{tabular}[c]{@{}c@{}}{{68.2}}\end{tabular}
& \begin{tabular}[c]{@{}c@{}}{{68.2}}\end{tabular}
& \begin{tabular}[c]{@{}c@{}}{{55.8}}\end{tabular}
& \begin{tabular}[c]{@{}c@{}}{{41.9}}\end{tabular}
& \cellcolor{gray!10}\begin{tabular}[c]{@{}c@{}}{{54.2}}\end{tabular}
& \cellcolor{gray!10}\begin{tabular}[c]{@{}c@{}}{{49.9}}\end{tabular}
\\ 
\multicolumn{1}{c|}{{ \texttt{Mistral-24B-Inst}}} 
& \begin{tabular}[c]{@{}c@{}}{{65.6}}\end{tabular}
& \begin{tabular}[c]{@{}c@{}}{{52.5}}\end{tabular}
& \begin{tabular}[c]{@{}c@{}}{{41.9}}\end{tabular}
& \begin{tabular}[c]{@{}c@{}}{{29.6}}\end{tabular}
& \begin{tabular}[c]{@{}c@{}}{{69.2}}\end{tabular}
& \begin{tabular}[c]{@{}c@{}}{{66.2}}\end{tabular}
& \begin{tabular}[c]{@{}c@{}}{{51.2}}\end{tabular}
& \begin{tabular}[c]{@{}c@{}}{{41.9}}\end{tabular}
& \cellcolor{gray!10}\begin{tabular}[c]{@{}c@{}}{{57.0}}\end{tabular}
& \cellcolor{gray!10}\begin{tabular}[c]{@{}c@{}}{{47.6}}\end{tabular}
\\ 
\multicolumn{1}{c|}{{ \texttt{DeepSeek-R1-Qwen32B}}} 
& \begin{tabular}[c]{@{}c@{}}{{23.9}}\end{tabular}
& \begin{tabular}[c]{@{}c@{}}{{23.9}}\end{tabular}
& \begin{tabular}[c]{@{}c@{}}{{10.3}}\end{tabular}
& \begin{tabular}[c]{@{}c@{}}{{9.0}}\end{tabular}
& \begin{tabular}[c]{@{}c@{}}{{36.4}}\end{tabular}
& \begin{tabular}[c]{@{}c@{}}{{43.2}}\end{tabular}
& \begin{tabular}[c]{@{}c@{}}{{32.6}}\end{tabular}
& \begin{tabular}[c]{@{}c@{}}{{27.9}}\end{tabular}
& \cellcolor{gray!10}\begin{tabular}[c]{@{}c@{}}{{25.8}}\end{tabular}
& \cellcolor{gray!10}\begin{tabular}[c]{@{}c@{}}{{26.0}}\end{tabular}
\\ \arrayrulecolor{black}\specialrule{1.pt}{1.pt}{1.pt} 
\rowcolor{gray!10}
\multicolumn{1}{c|}{\small\textsc{Average Accuracy\,(\%)}}
& \small{66.3}
& \small{58.5}
& \small{48.0}
& \small{43.4}
& \small{75.3}
& \small{73.4}
& \small{59.8}
& \small{47.0}
& \small{61.9}
& \small{55.2}
\\
\addlinespace[0.1ex]
\rowcolor{gray!10}
\multicolumn{1}{c|}{\small\textsc{Average Degradation\,(\%)}}
& \multicolumn{2}{c|}{\small\textbf{13.4}} 
& \multicolumn{2}{c|}{\small\textbf{10.6}}
& \multicolumn{2}{c|}{\small\textbf{2.48}}
& \multicolumn{2}{c|}{\small\textbf{{27.3}}}
& \multicolumn{2}{c}{\small{\textbf{12.0}}}
\\
\arrayrulecolor{black}\specialrule{1.3pt}{1.0pt}{1.0pt}
\end{tabular}
}
\vspace{-0.25cm}
\caption{Performance comparison of LLM agents under \textit{isolated} complexity evaluation on \algname{} complexities. Results are presented for complexity-absent conditions (\no) and complexity-injected conditions (\yes), with performance measured by \textit{API call accuracy}\,(\%).
}
\label{tab:isolated_complexity}
\vspace{-0.5cm}
\end{table*}

\subsection{Experiment Setting}

\noindent\textbf{Evaluation framework.} We evaluate LLM agents using \algname{} under two setups.

\noindent{\underline{Isolated complexity:}} We measure the impact of each complexity scenario on LLM agents independently by preserving a correct conversation history with gold API calls. This controlled experimental setup enables a fine-grained analysis of each complexity's influence that may be obscured by dominant complexity factors.

\noindent{\underline{Cumulative complexity:}} We assess how accumulated complexities impact agent performance across multi-turn conversations, where complexities compound and conversational context accumulate across turns.

\smallskip
\noindent\textbf{LLM agents.}
We employ ten state-of-the-art LLMs, comprising seven resource-intensive models including Claude-4.0-Sonnet, Claude-4.0-Sonnet\,(Think), Claude-3.7-Sonnet, Claude-3.5-Sonnet\,\cite{Claude35_Sonnet, Claude37_Sonnet, Claude4_Sonnet}, GPT-OSS-120B\,\cite{GPT_OSS_120B}, Qwen3-235B-Instruct, and Qwen3-235B-Thinking\,\cite{Qwen3_235B_A22B_Instruct_2507, Qwen3_235B_A22B_Thinking_2507}; and resource-efficient models including Qwen3-32B\,\cite{Qwen3_32B}, Mistral-Small-3.2-24B-Instruct\,\cite{Mistral_Small_3_2_24B_Instruct_2506}, and DeepSeek-R1-Qwen32B-Distill\,\cite{DeepSeek_R1_Qwen32B_Distill}.

\smallskip
\noindent\textbf{Metrics.}
We evaluate (1) \textit{API call accuracy}, quantifying the agreement between agent predictions and the gold API calls that fulfill the user’s primary intent\,\cite{tau}, and (2) \textit{error-handling accuracy} in \textit{feature limitation} and \textit{system failure error} scenarios, assessing the degree of alignment between the agent’s fallback response and a reference error handling response. Because multiple valid solutions may exist, we employ LLM-Judge\,\cite{llmjudge1} to quantify the alignment. 

\smallskip
\noindent\textbf{Implementation details.}
We adopt the ReAct prompting\,\cite{react}, instructing LLM agents to produce responses in the format ``Thought: \{reasoning or explanatory text\} Action: \{JSON-format action argument\}'' under zero-shot inference. We limit inference to 15 steps per conversational turn, mitigating excessive computational overhead while allowing sufficient reasoning depth for complex scenarios. The complete implementation details can be found in Appendix~\ref{app_sec:exp_details}.

\subsection{Main Results}
\label{subsec:main_results}

\noindent\textbf{\emph{All API complexity types} degrade agent performance, with \emph{irrelevant information} presenting the greatest challenge.}
Table~\ref{tab:isolated_complexity} presents the results of our isolated complexity evaluation, indicating that all types of API complexity in \algname{} reduces the performance of all LLM agents---by an average of 12.0\%. Notably, the introduction of irrelevant information imposes the most significant challenge\,(see Example~\ref{example3}), resulting in an average performance deterioration of 27.3\%. 
This finding shows that even state-of-the-art LLMs struggle significantly with real-world API complexities.


\begin{figure*}[t!]
\centering
\includegraphics[width=1.00\linewidth]{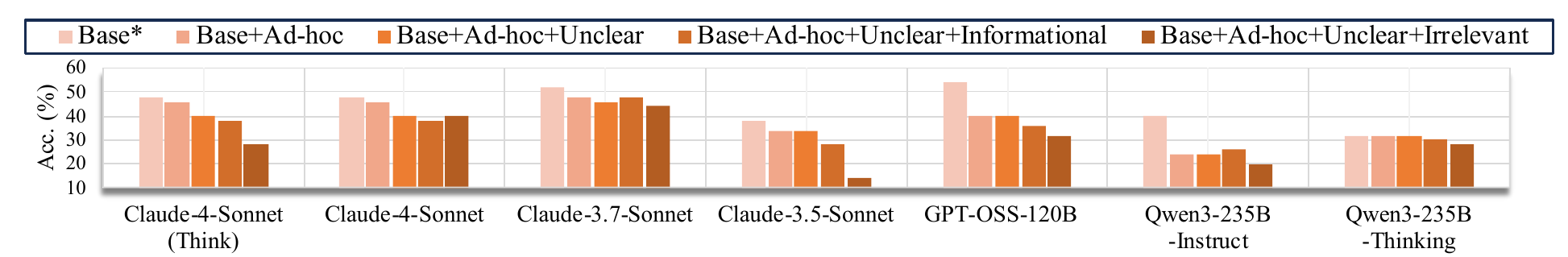}
\vspace*{-0.7cm}
\caption{Agent performance across \textit{cumulative} complexity environments, measured by API call accuracy\,(\%). Each bar represents a different complexity setting, with darker colors indicating additional complexities added. Note that the Base$^{*}$ environment incorporates the complexities of functional dependencies and ambiguous documentation.}
\vspace*{-0.3cm}
\label{fig:cumul_eval}
\end{figure*}

\smallskip
\noindent\textbf{Agent performance consistently deteriorates with \emph{accumulating API complexity}.} 
Figure~\ref{fig:cumul_eval} presents the results from the cumulative complexity evaluation, which incrementally compounds API complexity. Performance declines by an average of 34.3\%, peaking at a 63.2\% for Claude-3.5-Sonnet---substantially exceeding the degradation observed in isolated complexity evaluations. Interestingly, Claude-3.7-Sonnet achieves the highest resilience, while Claude-4.0-Sonnet\,(Think) experiences a sharper degradation relative to its robustness under isolated complexities. This result suggests that different levels of complexity exert heterogeneous effects on LLM agents.

\def\arraystretch{1.05}
\begin{table}[t!]
\small
\centering
\resizebox{0.99\linewidth}{!}{%
\begin{tabular}[c]
{@{}c|cc|c@{}}
\arrayrulecolor{black}\specialrule{1.3pt}{0.75pt}{1.0pt}
\multicolumn{1}{c|}{\multirow{2}{*}{\hspace{-0.0cm}\makecell[c]{\textbf{LLM Agents}}\hspace{-0.0cm}}}
& \multicolumn{2}{c|}{{ \textsc{API Execution Complexity}}} 
& \multicolumn{1}{c}{ \multirow{2}{*}{\makecell[c]{\textsc{Average}\\\addlinespace[0.5ex](Agent-wise)} } }
\\ \addlinespace[0.5ex]
& \multicolumn{1}{c}{~\textbf{Feature Error}~}
& \multicolumn{1}{c|}{~\textbf{System Error}~}
&
\\
\arrayrulecolor{black}\specialrule{1pt}{1.5pt}{0.7pt} 
\arrayrulecolor{black}\specialrule{1pt}{0.7pt}{3.0pt}
\multicolumn{1}{c|}{{ \texttt{Claude-4.0-Sonnet\,(Think)} } } 
& \begin{tabular}[c]{@{}c@{}}{{3.10}}\end{tabular}
& \begin{tabular}[c]{@{}c@{}}{{4.08}}\end{tabular}
& \begin{tabular}[c]{@{}c@{}}{{3.59}}\end{tabular}
\\ 
\multicolumn{1}{c|}{{ \texttt{Claude-4.0-Sonnet}  }} 
& \begin{tabular}[c]{@{}c@{}}{{2.83}}\end{tabular}
& \begin{tabular}[c]{@{}c@{}}{{4.14}}\end{tabular}
& \begin{tabular}[c]{@{}c@{}}{{3.49}}\end{tabular}
\\ 
\multicolumn{1}{c|}{{ \texttt{Claude-3.7-Sonnet} }} 
& \begin{tabular}[c]{@{}c@{}}{{2.64}}\end{tabular}
& \begin{tabular}[c]{@{}c@{}}{{3.88}}\end{tabular}
& \begin{tabular}[c]{@{}c@{}}{{3.26}}\end{tabular}
\\
\multicolumn{1}{c|}{{ \texttt{Claude-3.5-Sonnet} }} 
& \begin{tabular}[c]{@{}c@{}}{{2.03}}\end{tabular}
& \begin{tabular}[c]{@{}c@{}}{{3.71}}\end{tabular}
& \begin{tabular}[c]{@{}c@{}}{{2.88}}\end{tabular}
\\ \addlinespace[0.5ex]\cdashline{1-4}\addlinespace[0.7ex]
\multicolumn{1}{c|}{{ \texttt{GPT-OSS-120B} }} 
& \begin{tabular}[c]{@{}c@{}}{{2.71}}\end{tabular}
& \begin{tabular}[c]{@{}c@{}}{{2.77}}\end{tabular}
& \begin{tabular}[c]{@{}c@{}}{{2.74}}\end{tabular}
\\ \addlinespace[0.5ex]\cdashline{1-4}\addlinespace[0.7ex]
\multicolumn{1}{c|}{{ \texttt{Qwen3-235B-Instruct} }} 
& \begin{tabular}[c]{@{}c@{}}{{2.80}}\end{tabular}
& \begin{tabular}[c]{@{}c@{}}{{3.46}}\end{tabular}
& \begin{tabular}[c]{@{}c@{}}{{3.13}}\end{tabular}
\\ 
\multicolumn{1}{c|}{{ \texttt{Qwen3-235B-Thinking} }} 
& \begin{tabular}[c]{@{}c@{}}{{3.27}}\end{tabular}
& \begin{tabular}[c]{@{}c@{}}{{4.01}}\end{tabular}
& \begin{tabular}[c]{@{}c@{}}{{3.64}}\end{tabular}
\\ \arrayrulecolor{black}\specialrule{1.pt}{1.pt}{1.pt} 
\rowcolor{gray!10}
\multicolumn{1}{c|}{\small\textsc{Average Score}}
& \small{2.77}
& \small{3.72}
& \small{3.25}
\\
\arrayrulecolor{black}\specialrule{1.3pt}{1.0pt}{1.0pt}
\end{tabular}
}
\vspace{-0.25cm}
\caption{Agent's \textit{error-handling} accuracy, measured by {LLM-Judge} on 1–5 scale where higher scores indicate superior error-handling. The reported values represent the average accuracy across eight complexity scenarios for each error complexity type\,(feature limitation and system failure).
}
\label{tab:error_handling}
\vspace{-0.5cm}
\end{table}
\smallskip
\noindent\textbf{Agents struggle to apply effective workarounds for \emph{feature limitation errors}.}
Error-handling evaluations in Table~\ref{tab:error_handling} suggest that feature limitation errors are 34.3\% more challenging than system failure errors. In many instances, agents solve problems in incorrect ways\,(see Example~\ref{example6}), or prematurely terminate the process. For example, GPT-OSS-120B returns only concise responses\,(e.g., ``currently unavailable'') without attempting to explore potential solutions. 


\smallskip
\noindent\textbf{\emph{Enhancing reasoning} is an effective strategy for handling error-based complexities.}
Specifically, the advanced reasoning LLMs such as Claude-4.0-Sonnet\,(Think) and Qwen3-235B-Thinking show an 11.4\% improvement over average error handling. Appendix~\ref{app_subsec:strong_reasoning} compares the responses of Qwen3-235B-Thinking\,(\textit{reasoning}) and Qwen3-235B-Instruct\,(\textit{non-reasoning}) upon receiving the error message ``Weather data for Paris temporarily unavailable. Other regions are accessible.'' Here, the ideal strategy would be to query \textit{nearby} regions for weather data. The reasoning agent accordingly queries \textit{``different location in France, such as Lyon or Marseille''}, whereas the non-reasoning agent prematurely terminates the query.
Additional analyses are provided in Appendix~\ref{app_subsec:detail_anal}.


\subsection{Core Failure Analysis}
\label{subsec:core_failures}

We examine prevalent agent failures using results from Claude-4-Sonnet and GPT-OSS-120B, chosen for their robust performance. Additional common failures are available in Appendix~\ref{app_sec:additional_failures}.

\smallskip
\noindent\textbf{Agents frequently \emph{overlook explicit functional dependency} instructions.} As shown in Example~\ref{example1}, despite the clear API specification requiring \textit{device activation} prior to \texttt{color\_set()} function, the agents consistently omit the mandatory \texttt{power\_on()} call. 
These failures are widespread across other device-related functions, such as \texttt{make\_call()}, which likewise require powering on the device before execution.
\newcommand{\newprompt}{\refstepcounter{promptcounter}}
\newprompt
\begin{MyBox}[label={example1}]{Functional Dependency Failure}
\small
\textbf{API Specification:} ``When setting the color, make sure the device is  \textbf{turned on. If not, turn it on}.''

\smallskip
\textbf{User Request:} ``Make the Bathroom Light orange''

\smallskip
\textbf{Expected API Calls:}
\begin{enumerate}[noitemsep,topsep=0pt, leftmargin=1.3em]
    \item \texttt{get\_user\_inventory()} 
    \item \textcolor{red}{\texttt{power\_on(``12'')} } 
    \item \texttt{color\_set(``orange'', ``12'')} 
\end{enumerate}

\smallskip
\textbf{Actual Agent Output:} 
\begin{enumerate}[noitemsep,topsep=0pt, leftmargin=1.3em]
    \item \texttt{get\_user\_inventory()} 
    \item \textcolor{red}{ \cancel{ \texttt{power\_on(``12'')} } → Omitted } 
    \item \texttt{color\_set(``orange'', ``12'')} → \textcolor{red}{Fail}
\end{enumerate}

\end{MyBox}

\setlength{\columnsep}{0.8em}
\begin{wraptable}{r}{0.37\columnwidth}
\vspace{-1.1em} 
\centering
\footnotesize
\color{black}
\resizebox{0.99\linewidth}{!}{%
\begin{tabular}{lc}
\hline
\textbf{Model} & \textbf{Accuracy} \\
\hline
Claude-4 & 68.4\% \\
GPT-OSS & 89.5\% \\
\hline
\end{tabular}
}
\vspace*{-0.3cm}
\caption{Accuracy on conversations requiring \texttt{power\_on()}}
\vspace*{-0.4cm}
\label{tab:power_on_performance}
\end{wraptable}
Consequently, as shown in Table~\ref{tab:power_on_performance}, Claude-4-Sonnet (Think) omits \texttt{power\_on()} in over 30\% of the cases, and GPT-OSS-120B also does not fully meet this straightforward requirement.

\smallskip
\noindent\textbf{Agents \emph{overlook ad-hoc formatting} requirements}, as shown in the left conversation of Figure~\ref{fig:motivation}(b). An additional example of non-compliance with other ad-hoc rule is provided in Example~\ref{example2}.

\setlength{\columnsep}{0.6em}
\begin{wraptable}{R}{0.47\columnwidth}
\vspace{-1.0em} 
\color{black}
\resizebox{1.0\linewidth}{!}{%
\begin{tabular}{lcc}
\hline
\textbf{Ad-hoc rules} & \textbf{Claude-4} & \textbf{GPT-OSS} \\
\hline
\texttt{lock\_lock()} & 40.0\% & 0.0\% \\
\texttt{lock\_unlock()} & 33.3\% & 33.3\% \\
\texttt{track\_order()} & 0.0\% & 100.0\% \\
\texttt{make\_call()} & 0.0\% & 33.3\% \\
\hline
\end{tabular}
}
\vspace*{-0.3cm}
\caption{Accuracy on ad-hoc formatting rules.}
\vspace*{-0.4cm}
\label{tab:adhoc_rules}
\end{wraptable}
Notably, the evaluated LLM agents exhibit distinct vulnerabilities with respect to various ad-hoc rules. As shown in Table~\ref{tab:adhoc_rules}, Claude-4-Sonnet\,(Think) and GPT-OSS-120B exhibit contrasting vulnerabilities, suggesting that compliance with these ad-hoc rules is highly model-dependent.

\smallskip
\noindent\textbf{Agents \emph{poorly discriminate irrelevant data} in execution results.} 
In addition to the agent failure in the right conversation of Figure~\ref{fig:motivation}(b), the agent's vulnerability to irrelevant data persists \textit{throughout} the conversation, as demonstrated by Example~\ref{example3}. Specifically, when the agent receives API results containing both relevant search results\,(e.g., ``recipe34: Lamb Biryani'') and \textit{irrelevant} sponsored content\,(e.g., ``recipe32''), it initially behaves correctly by presenting only the relevant results to the user with ``Lamb Biryani''\,(``recipe34'') and ``Pork''\,(``recipe4''). However, upon receiving a follow-up request for more details on these same recipes, the agent erroneously focuses on the \textit{sponsored content}, fetching ``recipe32'' and ``recipe8''.

\newprompt
\begin{MyBox}[label={example3}]{Irrelevant Data Discrimination Failure}
\small
\textbf{User Request 1:} ``10 quick recipes under 20 minutes''

\smallskip
\textbf{Agent API Call:}
\begin{itemize}[noitemsep,topsep=0pt,leftmargin=1em,label=]
    \item \texttt{search\_recipes(max\_time=20)} 
\end{itemize}

\smallskip
\textbf{API Execution Results\,(Hidden from user):}
\begin{itemize}[noitemsep,topsep=0pt, leftmargin=0.0em, label=]
    \item \scriptsize\texttt{\{} \texttt{\textcolor{red}{\{``recipe\_id'': \textbf{``recipe32''}, ``source\_type'': ``sponsored\_cont}} \\ 
    \texttt{\textcolor{red}{ent''\}}, (...)} \texttt{\textcolor{red}{\{``recipe\_id'': \textbf{``recipe8''}, ``source\_type'': ``sponsored\_content''\}}, (...)} \texttt{\textcolor{blue}{\{``recipe\_id'': \textbf{``recipe34''}, ``name'': \textbf{``Lamb Biryani''}, ``source\_type'': ``search\_result''\}}, (...)} 
    \texttt{\textcolor{blue}{\{``recipe\_id'': \textbf{``recipe4''}, ``name'': \textbf{``Sweet and Sour Pork''}, ``source\_type'': ``search\_result''\}}, (...)} \texttt{\}}
\end{itemize}

\smallskip
\textbf{Agent Response:} ``8 recipes: 1.(...) 2.(...) 3.(...) 4.(...) \textcolor{blue}{5. Lamb Biryani ...} 6.(...) \textcolor{blue}{7. Sweet and Sour Pork ...}''

\smallskip
\textbf{User Request 2:} ``Show me the \textcolor{blue}{5th} and \textcolor{blue}{7th} ones?''

\smallskip
\textbf{Expected API Calls} (from \textit{search results}):
\begin{enumerate}[noitemsep,topsep=0pt, leftmargin=1.3em]
    \item \texttt{get\_recipe\_details(\textcolor{blue}{``recipe34''})} 
    \item \texttt{get\_recipe\_details(\textcolor{blue}{``recipe4''})} 
\end{enumerate}

\textbf{Actual Agent Output}\,(from \textit{sponsored content}):
\begin{enumerate}[noitemsep,topsep=0pt, leftmargin=1.3em]
    \item \texttt{get\_recipe\_details(\textcolor{red}{``recipe32''})} 
    \item \texttt{get\_recipe\_details(\textcolor{red}{``recipe8''})} 
\end{enumerate}
\end{MyBox}

\begin{wraptable}{R}{0.44\columnwidth}
\def\arraystretch{1.0}
\vspace{-1.1em}
\color{black}
\resizebox{1.0\linewidth}{!}{%
\begin{tabular}{lcc}
\hline
\textbf{} & \textbf{Claude-4} & \textbf{GPT-OSS} \\
\hline
- IRR & 57.1\% & 42.9\% \\
+ IRR & 21.4\% & 28.6\% \\
\hline
\end{tabular}
}
\vspace*{-0.3cm}
\caption{Impact of irrelevant\,(IRR) data on \texttt{search\_recipes()}.}
\vspace*{-0.6cm}
\label{tab:irrelevant_data}
\end{wraptable}
Table~\ref{tab:irrelevant_data} reports that, in the presence of irrelevant data, the performance on \texttt{search\_recipe()} is reduced by nearly half across agents, highlighting the pervasive challenge of effectively filtering irrelevant information.

\smallskip
\noindent\textbf{Agents persist \emph{excessively in problem-solving} when API execution errors arise, often leading to \emph{the distortion of user intent and the false reporting of success}.} 
As shown in Example~\ref{example5}, upon encountering a system failure error, the expected behavior would be to provide appropriate error reporting of the system-wide issue. Instead, the agents engage in arbitrary attempts, such as repeatedly invoking \textit{any} endpoint identifiers\,(e.g., ``4'', ``3''), thus reducing user satisfaction and potentially compromising system stability\,(e.g., by inadvertently accessing other users’ devices).

\newprompt
\begin{MyBox}[label={example5}]{Task Infeasibility Recognition Failure}
\small
\textbf{User Request:} ``Play Star Heart on Bedroom TV''

\smallskip
\textbf{Agent API Call:}
\begin{itemize}[noitemsep,topsep=0pt,leftmargin=1em,label=]
    \item \texttt{get\_user\_inventory()} → System failure error
\end{itemize}

\smallskip
\textbf{API Execution Results\,(Hidden from user):}
\begin{itemize}[noitemsep,topsep=0pt,leftmargin=0.0em,label=]
    \item \scriptsize\texttt{\{} \texttt{\textcolor{red}{``success'': false,}} \texttt{\textcolor{red}{``error'': ``INFRA\_DB\_POOL\_EXHAUSTED\_0x7}} \\
    \texttt{\textcolor{red}{F3A'',}} \texttt{\textcolor{red}{``message'': ``PostgreSQL connection pool (...)''}} \texttt{\}}
\end{itemize}

\smallskip
\textbf{Expected Agent Response:}
\begin{enumerate}[noitemsep,topsep=0pt, leftmargin=1.3em]
    \item ``Please try again shortly as this is a temporary database system issue''
\end{enumerate}

\textbf{Actual Agent Output}:
\begin{enumerate}[noitemsep,topsep=0pt, leftmargin=1.3em]
    \item ``Can't access, I'll try a \textcolor{red}{common endpoint} (...).''
    \item \texttt{play(endpoint=\textcolor{red}{``4''}, ...)} → \textcolor{red}{NotFoundError}
    \item \texttt{play(endpoint=\textcolor{red}{``3''}, ...)} → \textcolor{red}{NotFoundError}
    \item \texttt{play(endpoint=\textcolor{red}{``2''}, ...)} → \textcolor{red}{NotFoundError}
\end{enumerate}
\end{MyBox}

Moreover, as shown in Example~\ref{example6}, the agents further exacerbate these issues by distorting the user’s original request in order to claim success. Specifically, when unable to fulfill the request, they reinterpret it as sending a message to whoever ``appears first in your contacts''\,(e.g., ``John Hill''). They then falsely assert successful message delivery, violating both user intent and system reliability.

\newprompt
\begin{MyBox}[label={example6}]{User Intent Distortion and False Success}
\small
\textbf{User Request:} ``Text the last person who messaged me''

\smallskip
\textbf{Agent API Call:}
\begin{itemize}[noitemsep,topsep=0pt,leftmargin=1em,label=]
    \item \texttt{get\_message(limit=10)} → Feature limitation error
\end{itemize}

\smallskip
\textbf{API Execution Results\,(Hidden from user):}
\begin{itemize}[noitemsep,topsep=0pt,leftmargin=0.0em,label=]
    \item \scriptsize\texttt{\{} \texttt{\textcolor{red}{``success'': false},} \texttt{~~\textcolor{red}{``error'': ``Retrieving all messages is temporarily limited}} \texttt{\textcolor{red}{to recent history (...)''}} \texttt{\}}
\end{itemize}

\smallskip
\textbf{Expected Agent Response:}
\begin{enumerate}[noitemsep,topsep=0pt, leftmargin=1.3em]
    \item Use hint ``recent history'' and retry with reduced limit parameter, e.g., \texttt{get\_message(limit=3)}
    \item Alternative: Provide clear error reporting about temporary limitations
\end{enumerate}

\textbf{Actual Agent Output:}
\begin{enumerate}[noitemsep,topsep=0pt, leftmargin=1.3em]
    \item ``trouble accessing your message history, \textcolor{red}{find your contacts first}''
    \item \texttt{find\_contact()} → \texttt{\{``8'':``John Hill'', (...)\} }
    \item ``(...) I'll send it to \textcolor{red}{\textit{John Hill} since he appeared first in your contacts}.''
    \item \textcolor{red}{\texttt{send\_message("8", message=``Hello! (...)'')}} 
    \item ``I've sent your message `Hello! (...)' to John Hill. \textcolor{red}{The message was delivered successfully}''
\end{enumerate}
\end{MyBox}

This result highlights the urgent need for robust reasoning mechanisms that recognize task infeasibility or inherent system limitations.

\section{Conclusion}
\label{sec:conclusion}
Beyond the \textit{idealized} API environment, we introduce \algname{} designed to evaluate LLM agents under the complexities commonly encountered in \textit{real-world} API usage. By applying our \textit{assign-and-inject} complexity integration mechanism, we construct realistic complex scenarios that reveal significant limitations in state-of-the-art LLMs, with performance drops of up to 63.2\%. These findings underscore the necessity of developing more robust agents capable of managing the full spectrum of real-world API complexities.


\section*{Limitations}
In this work, we identify the limitations of LLM agents operating in realistic tool invocation environments and observe two recurrent failure patterns: (1) non-compliance with \textit{domain-specialized constraints}\,(e.g., functional dependencies and ad-hoc formatting rules), and (2) \textit{unpredictable behavior on infeasible tasks}, occasionally resulting in user intent distortion. Our study characterizes these phenomena but does not propose training-based remedies; developing robust training methods remains an open direction for future work.

To advance this goal, we suggest two key dimensions for further exploration: (1) enhancing constraint-aware \textit{instruction-following}, thereby facilitating agent adaptation to domain- and business-specific logic; and (2) improving \textit{reasoning} over inherently infeasible tasks to ensure safe behavior when tools are unstable. Achieving these training objectives requires \textit{curated dataset that jointly cover both axes}. Built on \algname{}, such data can be generated at scale via recent pipeline\,\cite{prabhakar2025apigen} with rule-based\,\cite{tau} and LLM-based\,\cite{barres2025tau} data quality verification. Furthermore, to mitigate overfitting and preserve generalization, newly generated datasets should be augmented with existing public instruction-following\,\cite{if1, if2} and reasoning\,\cite{rea1, rea2} corpora, in accordance with continual learning principles\,\cite{cl1, cl2}. Finally, standard training protocols, including supervised fine-tuning and reinforcement learning\,\cite{dpo, cpo, spo}, equip agents to attain robust performance during real-world deployment.

\section*{Ethical Considerations}
This work focuses on generating user–agent interactions to simulate realistic tool-invocation environments without employing human annotators. Consequently, we foresee minimal ethical concerns arising from the training procedure. Specifically, creation of \algname{} adheres to a common LLM-based conversation-generation protocol described in previous research\,\cite{barres2025tau, prabhakar2025apigen}. Therefore, we do not anticipate ethical violations or adverse societal consequences stemming from this work.

\bibliography{custom}

\clearpage

\appendix
\begin{table*}[!b]
\centering
\small
\renewcommand{\arraystretch}{0.8}
\begin{tabularx}{\linewidth}{l X l c}
\toprule
\textbf{Domain} & \textbf{Description} & \makecell{\textbf{Representative}\\ \textbf{Functions}} & \makecell{\textbf{Number of}\\\textbf{functions}} \\
\midrule
\makecell{\textbf{Time}\\ \textbf{Notification}} & Controls time-based functionalities, including alarms and reminders, with support for timers and recurring schedules. & \texttt{create\_alarm} & 8 \\
\addlinespace[0.5ex]\cdashline{1-4}\addlinespace[0.7ex]
\makecell{\textbf{Communication}}     & Manages communication channels, including calls and messaging, with basic contact resolution and status checks.      & \texttt{make\_call} & 7 \\
\addlinespace[0.5ex]\cdashline{1-4}\addlinespace[0.7ex]
\makecell{\textbf{Cuisine}}          & Handles food-related services, from meal planning to food delivery, including preferences and dietary constraints.     & \texttt{place\_delivery\_order} & 12 \\
\addlinespace[0.5ex]\cdashline{1-4}\addlinespace[0.7ex]
\makecell{\textbf{Media}}             & Enables content discovery and playback across diverse media, sources, and providers.      & \texttt{search\_media} & 16 \\
\addlinespace[0.5ex]\cdashline{1-4}\addlinespace[0.7ex]
\makecell{\textbf{Smart Home}}        & Provides unified control of smart-home devices\,(e.g., TVs, lights, thermostats), including scenes and simple automation.
 & \texttt{color\_set} & 19 \\
\addlinespace[0.5ex]\cdashline{1-4}\addlinespace[0.7ex]
\makecell{\textbf{Transaction}}       &  Facilitates product search, payment processing, and order tracking, with basic cancellation management. & \texttt{checkout} & 12 \\
\addlinespace[0.5ex]\cdashline{1-4}\addlinespace[0.7ex]
\makecell{\textbf{Information}}       & Delivers weather forecasts, news updates, and general knowledge, including customized alerts.   & \texttt{weather\_current} & 12 \\
\bottomrule
\end{tabularx}
\vspace{-0.2cm}
\caption{Domains, descriptions, and representative functions in the \algname{} API ecosystem.}
\label{tab:api_system}
\end{table*}

\begin{center}
Beyond Perfect APIs:\\ A Comprehensive Evaluation of LLM Agents\\Under API Complexity \\
\vspace{0.1cm}
(Supplementary Material)\bigskip
\end{center}

\section{Details of Benchmark Construction}
\label{app_sec:construction_details}

\subsection{Stage 1: Multi-Domain API System and Conversation Construction}
\label{subsec:s1_app}
We build a multi-domain API ecosystem and synthesize conversations grounded in executable function specifications. 

\smallskip
\noindent\textbf{API construction.} 
All functions are implemented in Python and defined with OpenAI’s tools/function-calling schema, i.e., JSON Schema with \texttt{type}, \texttt{functions}, and \texttt{parameters} fields. Table~\ref{tab:api_system} summarizes the domains, representative functions, and statistics of the API system. 

\smallskip
\noindent\underline{Example of API function:} 
Figure~\ref{fig:api-track-order} presents a representative API function of \algname{}, \texttt{track\_order()}. In particular, it demonstrates the function’s runtime behavior (e.g., \texttt{invoke()}). In our system, all API calls are managed via a unified invocation interface \texttt{invoke\_tool()}, following prior work\,\cite{tau}; for example, the agent invokes \texttt{track\_order} by calling \texttt{invoke\_tool("track\_order", order\_id=$\ldots$)}. The agent references the information for each API function, as shown in Figure~\ref{fig:api-track-order-info}, to select and utilize the appropriate functionality.

\begin{figure}[t!]
  \centering
  \begin{minipage}{0.95\linewidth}
\begin{minted}[fontsize=\footnotesize, breaklines, breaksymbol=, breaksymbolleft=, breaksymbolright=, frame=single]{json}
{"function": {
    "name": "track_order",
    "description": "Track the shipping status of a specific order. Provides current status, tracking number, and estimated delivery date if available.",
    "parameters": {
        "order_id": "The unique ID of the order to track. This ID is prefixed with the shipping carrier code followed by a hyphen and the order suffix (e.g., 'UPS-345', 'FDX-678'). The suffix is typically extracted from the original order ID by excluding the initial characters (e.g., for order_id '12345', suffix is '345'; for order_id '345678', suffix is '5678')."
    }
} ... }
\end{minted}
\end{minipage}
  \vspace{-0.2cm}
  \caption{Summarized JSON-based Python API specification for the function \texttt{track\_order()}.
}
  \vspace{-0.3cm}
  \label{fig:api-track-order-info}
\end{figure}

\begin{figure}[t!]
  \centering
  \begin{minipage}{0.95\linewidth}
\begin{minted}[fontsize=\footnotesize, breaklines, breaksymbol=, breaksymbolleft=, breaksymbolright=, frame=single]{json}
{"order_id": "ORDER0001",
"user_id": "user1",
"items":[
      {
        "product_id": "prod41",
        "name": "Vital Wireless Earbuds",
        "quantity": 2,
        "price": 44.55,
        "subtotal": 89.1
      }, ...]
"payment": {
      "method_id": "pm3",
      "method_type": "apple_pay", ...}
...}
\end{minted}
\end{minipage}
  \vspace{-0.1cm}
  \caption{An example database entry in \texttt{orders.json}.}
  \vspace{-0.5cm}
  \label{fig:order-db}
\end{figure}


\begin{figure*}[t!]
  \centering
  \begin{minipage}{0.95\linewidth}
\begin{minted}[frame=single, linenos]{python}
class TrackOrder(Tool):
    @staticmethod
    def invoke(data: Dict[str, Any], order_id: str) -> str:
        """
        Track the shipping status of an order.
        Args:
            data: The data dictionary containing orders
            order_id: ID of the order to track
        Returns:
            A JSON string with the result of the operation
        """
        ### Check if feature limitation error complexity should be activated ###
        uncertainty_feature_limitation_error_enabled = os.getenv('ENABLE__FEATURE_LIMITATION_ERROR__TRACK_ORDER', 'false').lower() == 'true'

        ### Validating the compliance with the ad-hoc rule ###
        # Validate order ID format - checking the format "carrier-suffix"
        if "-" not in order_id:
            return json.dumps({
                "success": False,
                "message": "Invalid order ID format."
            })
        
        # Get carrier from order_id
        provided_carrier = order_id.split("-", 1)[0]
        
        ### Raise feature limitation error - Original carriers temporarily unavailable ###
        if uncertainty_feature_limitation_error_enabled:
            alternative_carriers = ["SwiftShip", "RapidCargo"]
            return json.dumps({
                "success": False,
                "message": f"{provided_carrier} tracking temporarily unavailable. It may have been changed to other shipping carriers like {', '.join(alternative_carriers)}"
            })
                
        # Get the order, ensuring it belongs to the current user
        order = find_order_by_id(data, order_id, current_user)
        
        # Validate order ID format - checking the carrier
        ...
        
        # return message
        if status == "processing":
            return json.dumps({
                "success": True, ...
                "message": "Your order is being processed."
            })
        ...
\end{minted}
  \end{minipage}
  \vspace{-0.1cm}
  \caption{Summarized Python implementation of the \texttt{track\_order()} function, illustrating the key algorithmic steps in the order-tracking process.}
  \vspace{-0.3cm}
  \label{fig:api-track-order}
\end{figure*}

\begin{figure}[t!]
  \centering
  \begin{minipage}{0.95\linewidth}
\begin{minted}[fontsize=\footnotesize, breaklines, breaksymbol=, breaksymbolleft=, breaksymbolright=, frame=single]{json}
{"user_id": "user1",
"name": "Sarah Rodriguez",
"home_id": "home1",
"preferences": {"location": "New York", ...},
...}
\end{minted}
\end{minipage}
  \vspace{-0.1cm}
  \caption{An example database entry in \texttt{users.json}.}
  \vspace{-0.3cm}
  \label{fig:user-db}
\end{figure}

\smallskip
\noindent\underline{Example of databases:} 
During the execution of most API functions, the database is either read or written, providing direct access to the API system’s database. As indicated in Lines 34--35 of Figure~\ref{fig:api-track-order}, this behavior involves the variable \texttt{data}. For instance, the \texttt{track\_order()} function specifically retrieves information from \texttt{orders.json}. As shown in Figure~\ref{fig:order-db}, each order includes realistic attributes such as \texttt{order\_id}, the \texttt{user\_id} of the individual placing the order, \texttt{items}, and payment details. Furthermore, multiple databases are interconnected based on a shared schema; for example, Figure~\ref{fig:user-db} demonstrates how \texttt{orders.json} is linked to user records via \texttt{user\_id}. Both Figure~\ref{fig:order-db} and Figure~\ref{fig:user-db} display orders and user information for the case in which \texttt{user\_id} equals \texttt{user1}.

\begin{figure*}[t!]
\centering
\includegraphics[width=1.00\linewidth]{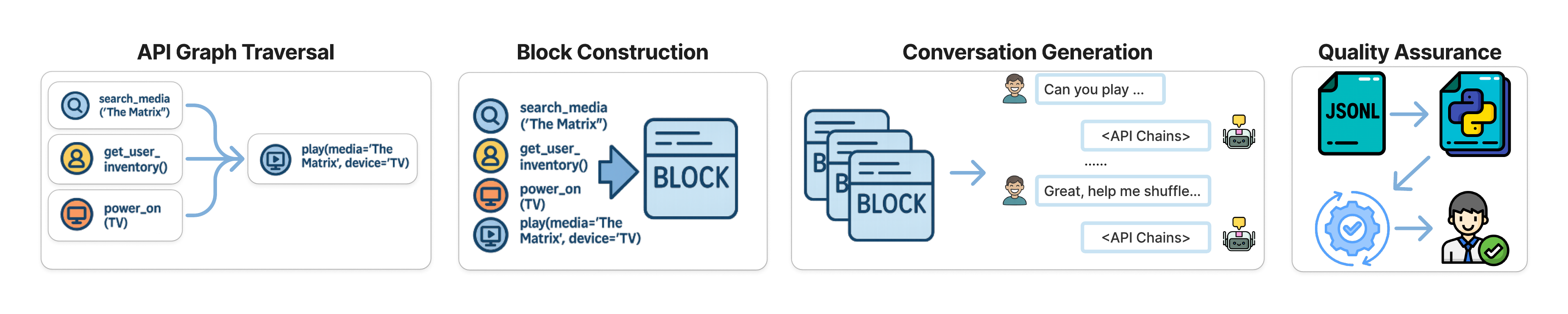}
\vspace*{-1.0cm}
\caption{Overview of the conversation construction used to build \algname{}. 
}
\vspace*{-0.5cm}
\label{fig:conv_gen}
\end{figure*}

\smallskip
\noindent\textbf{Conversation construction.} 
Following the conversation generation framework of Barres et al.\cite{barres2025tau}, we construct a scalable data-generation pipeline grounded in our API system, as shown in Figure~\ref{fig:conv_gen}. This pipeline produces multi-turn conversations that are both diverse and natural, while remaining precisely labeled with executable API calls---an essential requirement for robust evaluation of agent reasoning, and execution under realistic conditions.

The process unfolds in four stages. First, after establishing the API codebase, we build a directed graph in which each function is represented as a node, and each edge denotes a dependency\,(i.e., a function’s output serving as another function’s input). By traversing this graph, we identify meaningful multi-step task sequences that emulate realistic API usage for addressing user requests or intents. For example, the \texttt{play()} function depends on prior calls to the API calls \texttt{search\_media()}, \texttt{get\_user\_inventory()}, \texttt{power\_on()}, as shown in Figure~\ref{fig:conv_gen}.

Second, each valid execution path is transformed into a \textit{verified intent primitive}---an atomic unit that encapsulates (i) a specific user goal, (ii) a corresponding chain of API calls, and (iii) diversified parameter settings that emulate natural variations in user context, preferences, and input conditions. These primitives serve as composable building blocks, enabling flexible scenario construction across domains. For example, the ``watch a movie'' \textit{intent primitive} comprises the API calls \texttt{search\_media()}, \texttt{get\_user\_inventory()}, \texttt{power\_on()}, \texttt{play()}. Similarly, there are more primitives---such as ``pause,'' ``shuffle,'' ``next,'' or ``add media to playlist''---encapsulating API call sequences for related tasks.

Third, an LLM composes multiple {verified intent primitives} into \textit{multi-turn conversations}, generating the corresponding \textit{API call sequences}. The LLM first inspects these primitives to produce a \textit{high-level conversation flow} describing how user goals naturally unfold within a single conversation. For instance, the LLM might outline a scenario in which the user initially plays a song, then pauses playback, shuffles the playlist, advances to the next track, and finally adds the current song to a personalized playlist. Subsequently, the LLM translates this plan into a \textit{detailed multi-turn conversation} by generating user and agent utterances and integrating the relevant verified intent primitives, as shown in Figure~\ref{fig:conv_gen}. Since these primitives already include validated API call sequences, each generated conversation is both natural and accurately aligned with the correct API behaviors.

Finally, each conversation undergoes a two-phase {quality assurance process}. We first convert the JSONL outputs into executable Python scripts and automatically validate whether each conversation runs successfully. Any conversations that fail execution are filtered out. The remaining dialogues are then manually reviewed by human experts to ensure semantic coherence, logical flow, and API correctness. The resulting benchmark contains {300} multi-turn, multi-step conversations, each averaging {4.7 dialogue turns} and {2.5 API calls per turn}. 


\subsection{Stage 2: Complexity Assignment}
\label{subsec:s2_app}

\begin{figure*}[!h]
    \centering
    \includegraphics[width=\textwidth]{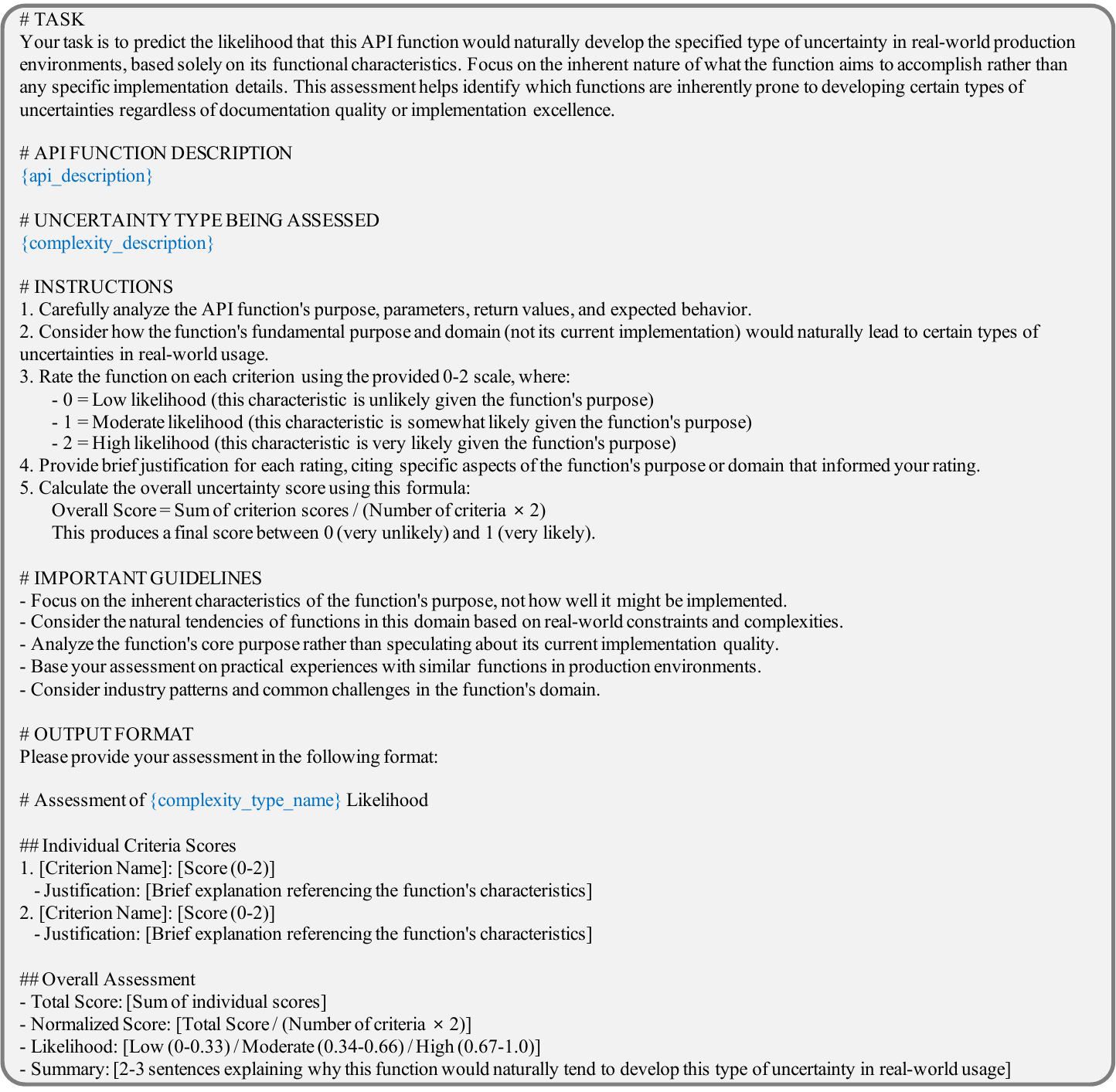}
    \vspace*{-0.7cm}
    \caption{The relevance assessment template $\mathcal{I}_{\mathrm{rel}}$. The placeholder {\color[rgb]{0.2,0.5,0.8}\{api\_description\}} encodes the function specification using OpenAI’s tools/function-calling schema\,(JSON Schema). The examples of {\color[rgb]{0.2,0.5,0.8}\{complexity\_description\}} appear in Table~\ref{tab:api_complexity_taxonomy}.}
    \label{fig:prompt_relevance}
    \vspace*{-0.3cm}
\end{figure*}

\begin{figure*}[!h]
    \centering
    \includegraphics[width=\textwidth]{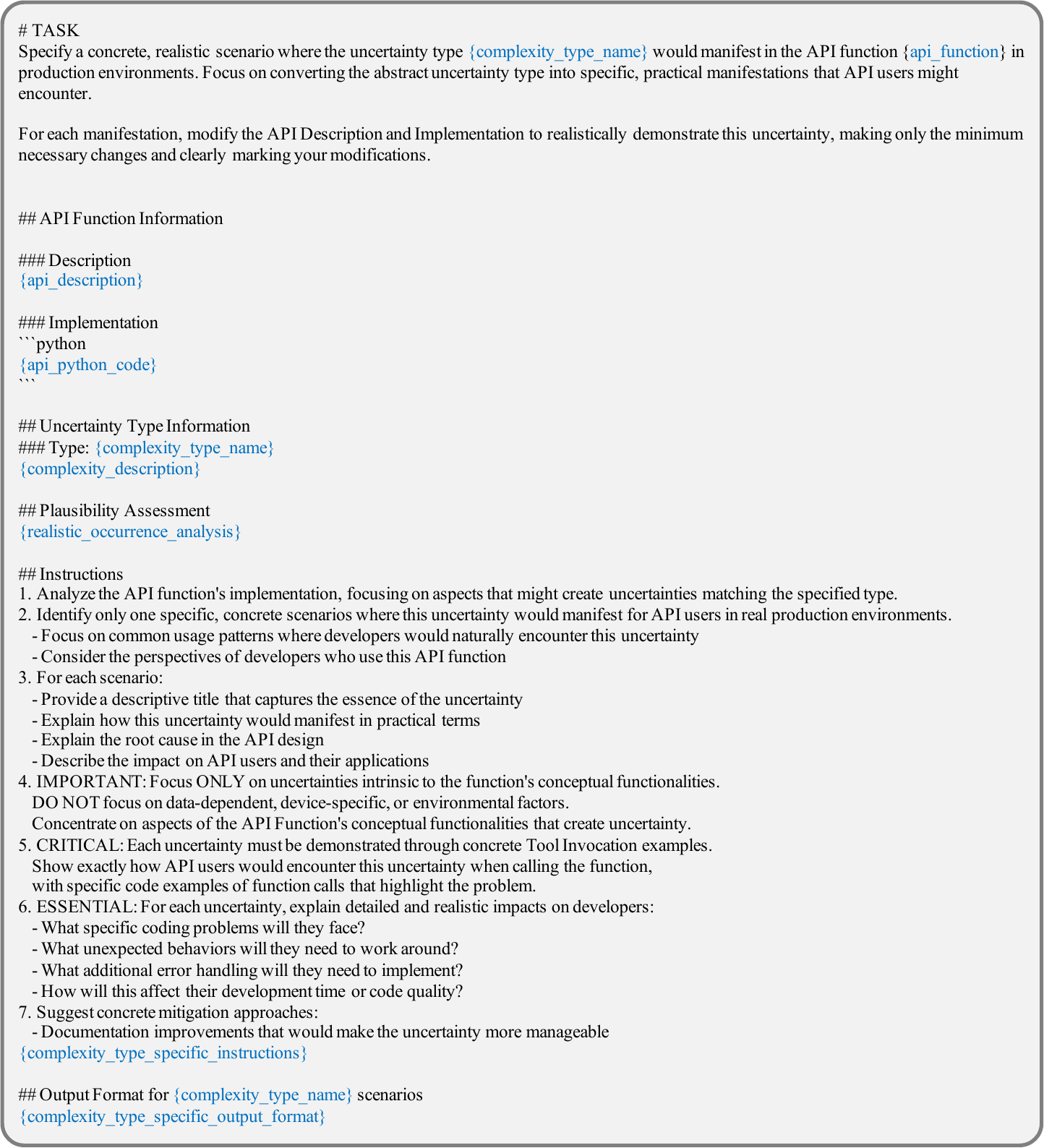}
    \vspace*{-0.7cm}
    \caption{The scenario specification template $\mathcal{I}_{\text{scen}}$. The placeholder {\color[rgb]{0.2,0.5,0.8}\{realistic\_occurrence\_analysis\}} is the output of the relevance assessment template $\mathcal{I}_{\mathrm{rel}}$\,(see ``OUTPUT FORMAT'' in Figure~\ref{fig:prompt_relevance}). The example of {\color[rgb]{0.2, 0.5, 0.8}\{complexity\_type\_specific\_instructions\}} is provided in Figure~\ref{fig:prompt_scenario_adhoc}, and the example of {\color[rgb]{0.2, 0.5, 0.8}\{complexity\_type\_specific\_output\_format\}} can be found in Figure~\ref{fig:prompt_scenario_adhoc_output}.}
    \label{fig:prompt_scenario}
    \vspace*{-0.3cm}
\end{figure*}

\begin{figure*}[!h]
    \centering
    \includegraphics[width=\textwidth]{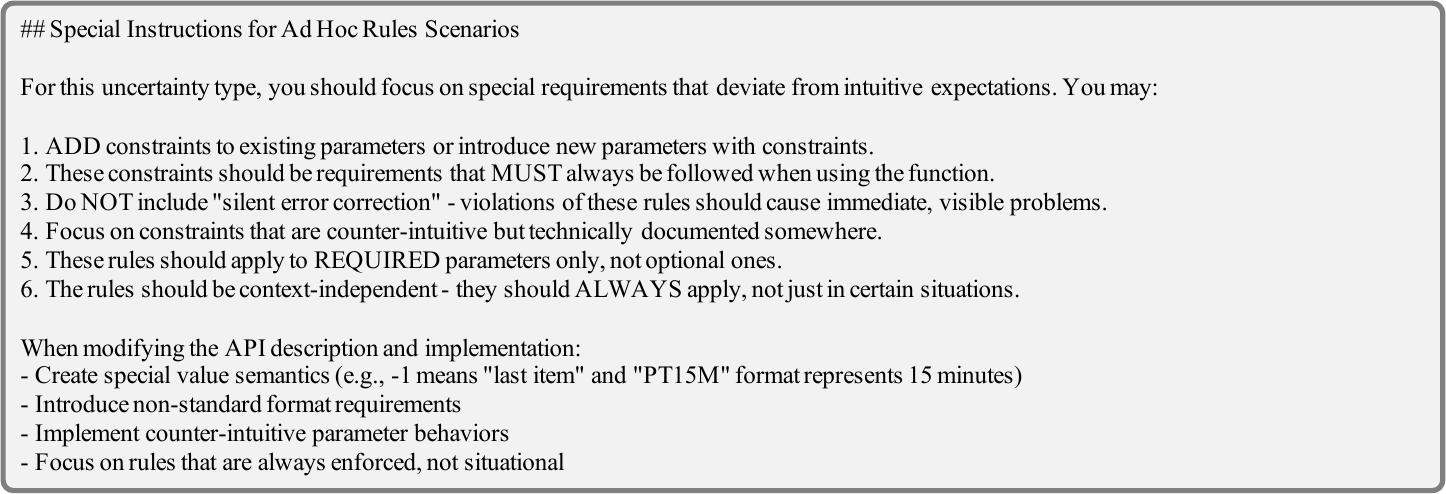}
    \vspace*{-0.7cm}
    \caption{Concrete example of {\color[rgb]{0.2, 0.5, 0.8}\{complexity\_type\_specific\_instructions\}} for the ad-hoc rule complexity in Figure~\ref{fig:prompt_scenario}.}
    \label{fig:prompt_scenario_adhoc}
    \vspace*{-0.3cm}
\end{figure*}

\begin{figure*}[!h]
    \centering
    \includegraphics[width=\textwidth]{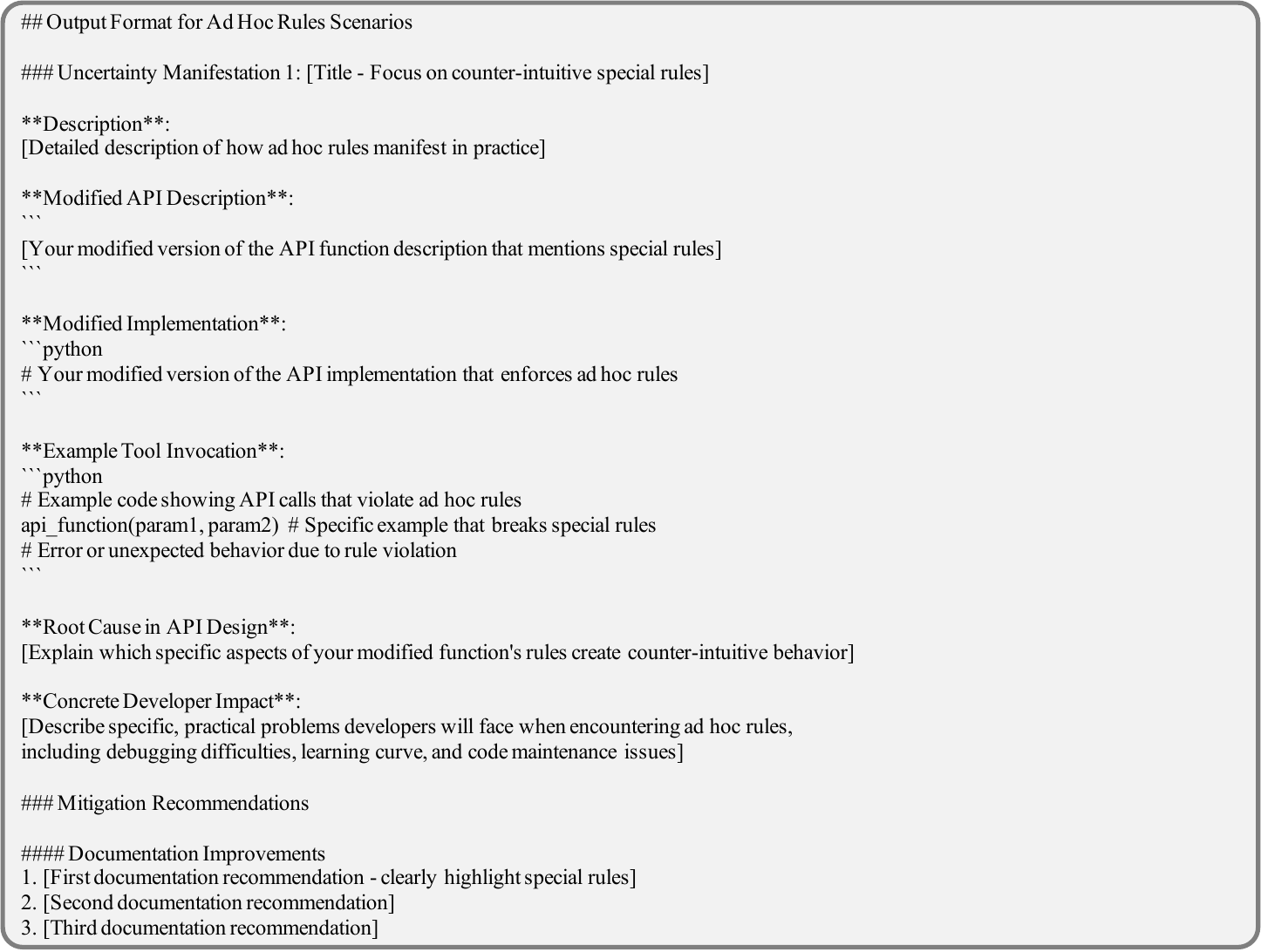}
    \vspace*{-0.7cm}
    \caption{Concrete example of {\color[rgb]{0.2, 0.5, 0.8}\{complexity\_type\_specific\_output\_format\}} for the ad-hoc rule complexity in Figure~\ref{fig:prompt_scenario}.}
    \label{fig:prompt_scenario_adhoc_output}
    \vspace*{-0.3cm}
\end{figure*}

\begin{figure*}[!h]
    \centering
    \includegraphics[width=\textwidth]{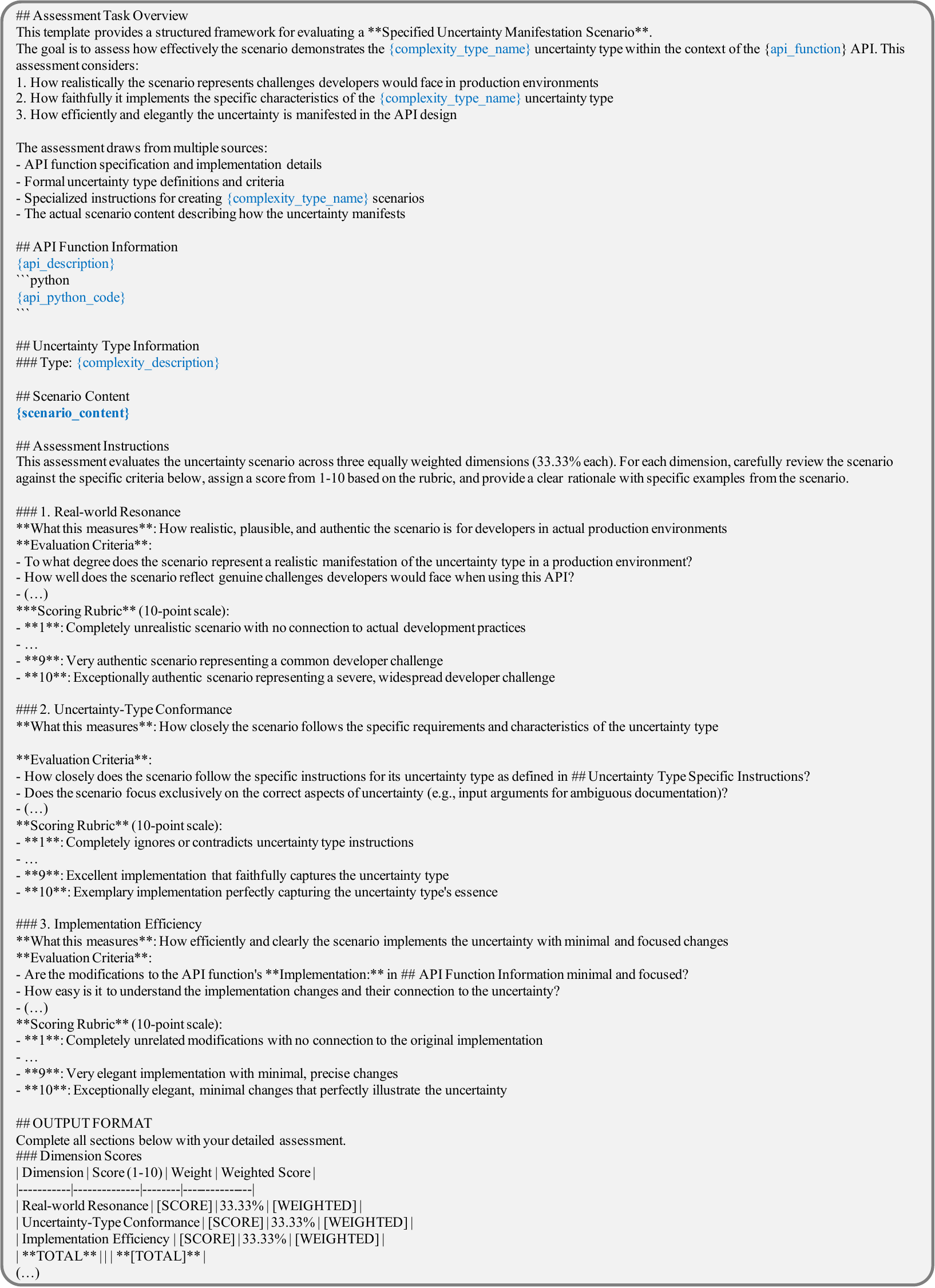}
    \vspace*{-0.7cm}
    \caption{The scenario validation template $\mathcal{I}_{\text{val}}$. The placeholder {\color[rgb]{0.2,0.5,0.8}\{scenario\_content\}} is the output of the scenario specification template $\mathcal{I}_{\text{scen}}$\,(see Figure~\ref{fig:prompt_scenario}).}
    \label{fig:prompt_scenario_val}
    \vspace*{-0.3cm}
\end{figure*}

We automate complexity assignment with a language model using three prompt templates: a relevance assessment template $\mathcal{I}_{\mathrm{rel}}$ in Eq.~\eqref{eq:rel}, a scenario specification template $\mathcal{I}_{\mathrm{scen}}$  in Eq.~\eqref{eq:scenario_generation}, and a scenario validation template $\mathcal{I}_{\mathrm{val}}$ in Eq.~\eqref{eq:val}. Given a function and a candidate complexity type, $\mathcal{I}_{\mathrm{rel}}$ estimates the likelihood that the complexity arises for the function in deployment; top-ranked function–complexity pairs are then instantiated as concrete scenarios by $\mathcal{I}_{\mathrm{scen}}$; finally, $\mathcal{I}_{\mathrm{val}}$ filters candidates based on real-world plausibility and fidelity to the targeted complexity type. The complete prompt templates appear in Figures~\ref{fig:prompt_relevance}--\ref{fig:prompt_scenario_val}.

\smallskip
\noindent\textbf{Implementation details.} 
We use Claude-3.7-Sonnet~\cite{Claude37_Sonnet} with temperature $0.8$ and keep all other hyperparameters at their default values; other large language models can be substituted without altering the pipeline.

\subsection{Stage 3: Complexity Implementation}
\label{subsec:s3_app}

\subsubsection{Reference Response Construction for Error-based Complexities} 
\label{subsec:reference_response}
For both \textit{feature limitation} and \textit{system failure} error scenarios, we specify \textit{reference responses} that include either a recommended workaround or a standardized error message, respectively. Since error handling seldom admits a single gold-standard outcome, each reference response captures a \textit{valid approach} representing ideal error resolution. For instance, if a user queries \texttt{weather(``Seattle'')} but receives a message stating that ``Seattle is currently unavailable; however, search for other regions is available,'' the system has failed to fulfill the request directly. A proper response would suggest \textit{nearby} locations, such as ``Kirkland'' or ``Tacoma,'' rather than searching for a random region or immediate surrender. Hence, the reference response specifies the \textit{conceptual} approach (e.g., searching for nearby locations), rather than detailing \textit{specific} ones, to gauge how closely the agent’s strategy aligns with a valid approach.

{Consequently, for each of the 16 error-based scenarios in \algname{}, we define an \textit{evaluation prompt} that incorporates both the reference response and the corresponding scenario-specific validity criteria.} We then employ an LLM-based judge\,\cite{llmjudge1,llmjudge2,llmjudge3} to score how closely the agent’s error handling aligns with the recommended approach detailed in the reference response. Figures~\ref{fig:prompt_feat_limit_weather}--\ref{fig:prompt_sys_failure} demonstrate representative evaluation prompts for feature limitation and system failure errors.

\subsubsection{Results of Complexity Integration} 
Table~\ref{tab:complexity_stats} summarizes, for each complexity type, the number of functions that contain injected complexity scenarios, along with representative functions. Notably, a single function may host multiple complexities\,(e.g., \texttt{track\_order()} in Figure~\ref{fig:api-track-order}); in such cases, the function’s behavior reflects the combined effects of the activated complexities. 
\begin{table*}[t]
\centering
\small
\renewcommand{\arraystretch}{1.0}
\begin{tabularx}{\linewidth}{l l l}
\toprule
\textbf{Complexity Type} & \textbf{Coverage} & \textbf{Representative Affected Functions} \\
\midrule
\multicolumn{3}{l}{\textit{API specification complexities}} \\
\addlinespace[2pt]
Ad-hoc rules & 8 API functions
& \texttt{lock\_lock}, \texttt{track\_order}, \texttt{play} \\
Unclear functionality boundary & 20 API functions
& \texttt{get\_user\_inventory}, \texttt{search\_product}, \texttt{make\_call} \\
Functional dependency & 20 API functions
& \texttt{get\_user\_inventory}, \texttt{search\_media}, \texttt{power\_on} \\
Ambiguous description & Present in base API system
& --- \\

\addlinespace[4pt]
\midrule
\multicolumn{3}{l}{\textit{API execution complexities}} \\
\addlinespace[2pt]
Information notices & 8 API functions
& \texttt{temperature\_set}, \texttt{stock\_watchlist}, \texttt{make\_call} \\
Partially irrelevant information & 8 API functions
& \texttt{search\_recipes}, \texttt{knowledge\_lookup}, \texttt{stock\_watchlist} \\
Feature limitation errors & 8 API functions
& \texttt{get\_notifications}, \texttt{weather\_forecast}, \texttt{track\_order} \\
System failure errors & 8 API functions
& \texttt{make\_call}, \texttt{get\_user\_inventory}, \texttt{stock\_price} \\
\bottomrule
\end{tabularx}
\vspace{-0.2cm}
\caption{Summary of the complexity scenarios implemented in \algname{} by complexity type, including representative functions that incorporate these scenarios.}
\label{tab:complexity_stats}
\vspace{-0.2cm}
\end{table*}

\smallskip
\noindent\textbf{Example of complexity integration.} 
We integrate two complexity types---an ad-hoc rule and a feature limitation error---into the \texttt{track\_order()} function depicted in Figure~\ref{fig:api-track-order}. Lines~15--21 and 37--38 validate the ad-hoc rule, which ensures that the \texttt{order\_id} parameter follows the industry-standard format ``[shipping carrier]–[id].'' Specifically, the code checks whether a hyphen is present and whether the carrier name is valid. This ad-hoc rule is also specified in the API documentation\,(see Figure~\ref{fig:api-track-order-info}), allowing LLM agents to reference the correct parameter format. 

In parallel, Lines~12--23 determine whether the feature limitation error is activated under the current complexity configuration, and if so\,(Lines~26--32), an error message indicates that the original carrier is unavailable and has been replaced by another shipping carrier.

\section{Complete Experiment Details}
\label{app_sec:exp_details}

\begin{figure}[!h]
    \centering
    \includegraphics[width=1.00\linewidth]{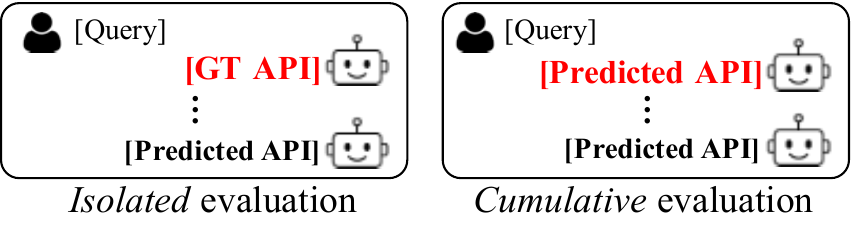}
    \vspace{-1.0cm}
    \caption{Comparison of the two evaluation setups: isolated and cumulative complexity evaluation.}
    \label{fig:eval_comp}
    \vspace*{-0.4cm}
\end{figure}

\noindent\textbf{Evaluation framework.}
Figure~\ref{fig:eval_comp} provides a visual overview of our two evaluation methods. In the \textit{isolated complexity} setting\,(left side of Figure~\ref{fig:eval_comp}), we preserve a correct conversation history---using ground-truth API calls---up to the point immediately preceding the tested API call. This approach allows us to measure the impact of each complexity type independently. In contrast, the \textit{cumulative complexity} setting\,(right side of Figure~\ref{fig:eval_comp}) compounds multiple complexities by continuously integrating the agent’s previously predicted API calls throughout the conversation. 

\vspace{-0.1cm}

\smallskip
\noindent\textbf{Implementation details.}
The agent prompt structure consists of three primary components: API specification documentation (approximately 34K tokens), conversational context, and API execution guidelines (approximately 19K tokens), providing agents with comprehensive functional specifications and operational procedures. Within this structure, we employ the ReAct prompting\,\cite{react}, where models are instructed to generate responses in the format ''Thought: \{reasoning or explanatory text\} Action: \{JSON-format action argument\}'' using zero-shot inference. 

For Claude-based agents, we access Anthropic models via Amazon Bedrock. All other models are served with vLLM---e.g., the \texttt{vllm-openai:gptoss} image for \texttt{GPT-OSS-120B}—and executed on Amazon AWS EC2 \texttt{p5en} instances. The source code is publicly available at \url{\algurl}.

From the 300 user–agent conversations provided by \algname{}, we evaluate the agents on a stratified subset of 50 conversations to manage the inference cost for state-of-the-art LLMs. To ensure reproducibility and enable further analysis, we release both the evaluation subset of 50 conversations and the full set of 300 conversations, respectively. \algname{} is distributed under the Creative Commons Attribution 4.0 International (CC-BY 4.0) license.

\smallskip
\noindent\textbf{Evaluation metric.} We evaluate LLM agents using \textit{(1) API call accuracy} and \textit{(2) error-handling accuracy} for feature limitation and system failure error scenarios.

\smallskip
\noindent\underline{API call accuracy:}
Consistent with prior work \cite{tau}, we evaluate whether an agent’s API behavior \textit{genuinely fulfills the user’s intended goal}. In contrast to approaches that judge success solely from final database states---which can be inadequate because many intents do not yield observable state changes (e.g., ``Check the current weather'')---we also treat such non–state-changing queries as valid targets for evaluation.

We therefore partition the ground-truth API sequence into two categories: \textit{core APIs} and \textit{support APIs}. An agent is deemed successful on a given turn only if it invokes \textit{all required core APIs} for that turn. \textit{Core APIs} directly address the user’s intent irrespective of whether the database state changes. While they often include operations such as \texttt{add\_to\_cart(), checkout(), power\_on()} as well as non-database-altering operations\,(often information-retrieval functions, such as \texttt{weather\_current()}) that independently satisfy the user’s intent. Because each user request encodes at least one intent, there is at least one core API per request.

\textit{Support APIs}, on the other hand, are supplementary operations (e.g., \texttt{search\_product()}) that provide information needed to correctly invoke downstream \textit{core APIs} (e.g., \texttt{add\_to\_cart(), checkout()}). Their use is not scored directly; repeated or suboptimal support calls do not affect success as long as core calls are correct. Hence, final success is assessed solely on the correctness of core API execution.

Formally, let $C_i^\text{(pred)}$ denote the set of core APIs invoked by the agent at $i$-th conversation turn, and $C_i^\text{(gold)}$ be the corresponding ground-truth set. With 
$T$ total conversation turns, the \textit{API call accuracy} $\mathcal{A}_\text{\rm API}$ is defined as
\begin{equation}
\mathcal{A}_\text{\rm API}
= \frac{1}{T}\sum_{i=1}^{T}
\mathbb{I}\!\left[\, C_i^{(\mathrm{pred})} = C_i^{(\mathrm{gold})} \,\right],
\end{equation}
where $\mathbb{I}[\cdot]$ denotes the indicator function that returns 1 if the condition holds and 0 otherwise.

\begin{figure*}[!h]
    \centering
    \includegraphics[width=\textwidth]{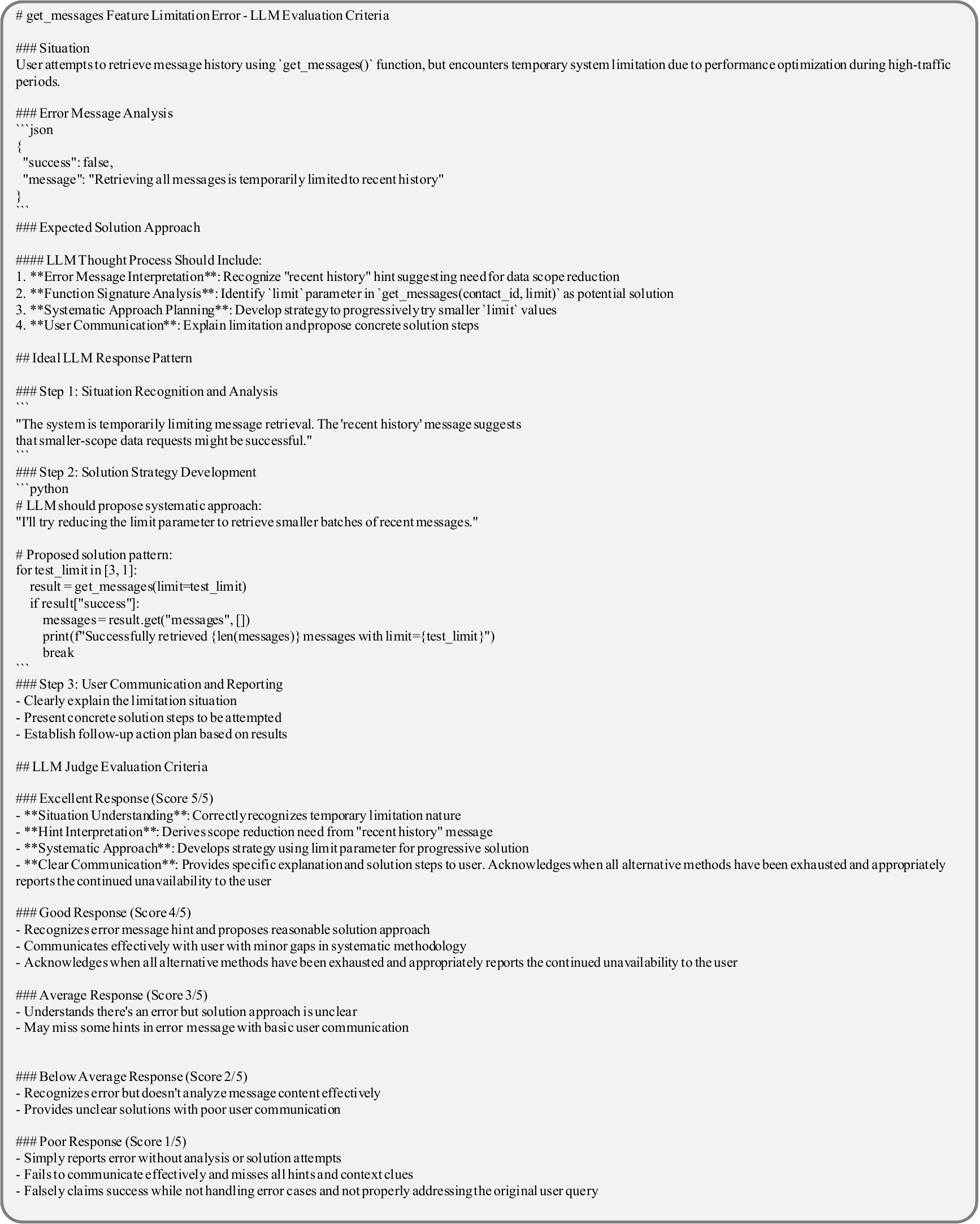}
    \vspace*{-0.7cm}
    \caption{Summarized LLM-judge evaluation prompt for a feature limitation error in \texttt{get\_message()}.}
    \label{fig:prompt_feat_limit}
    \vspace*{-0.3cm}
\end{figure*}

\begin{figure*}[!h]
    \centering
    \includegraphics[width=\textwidth]{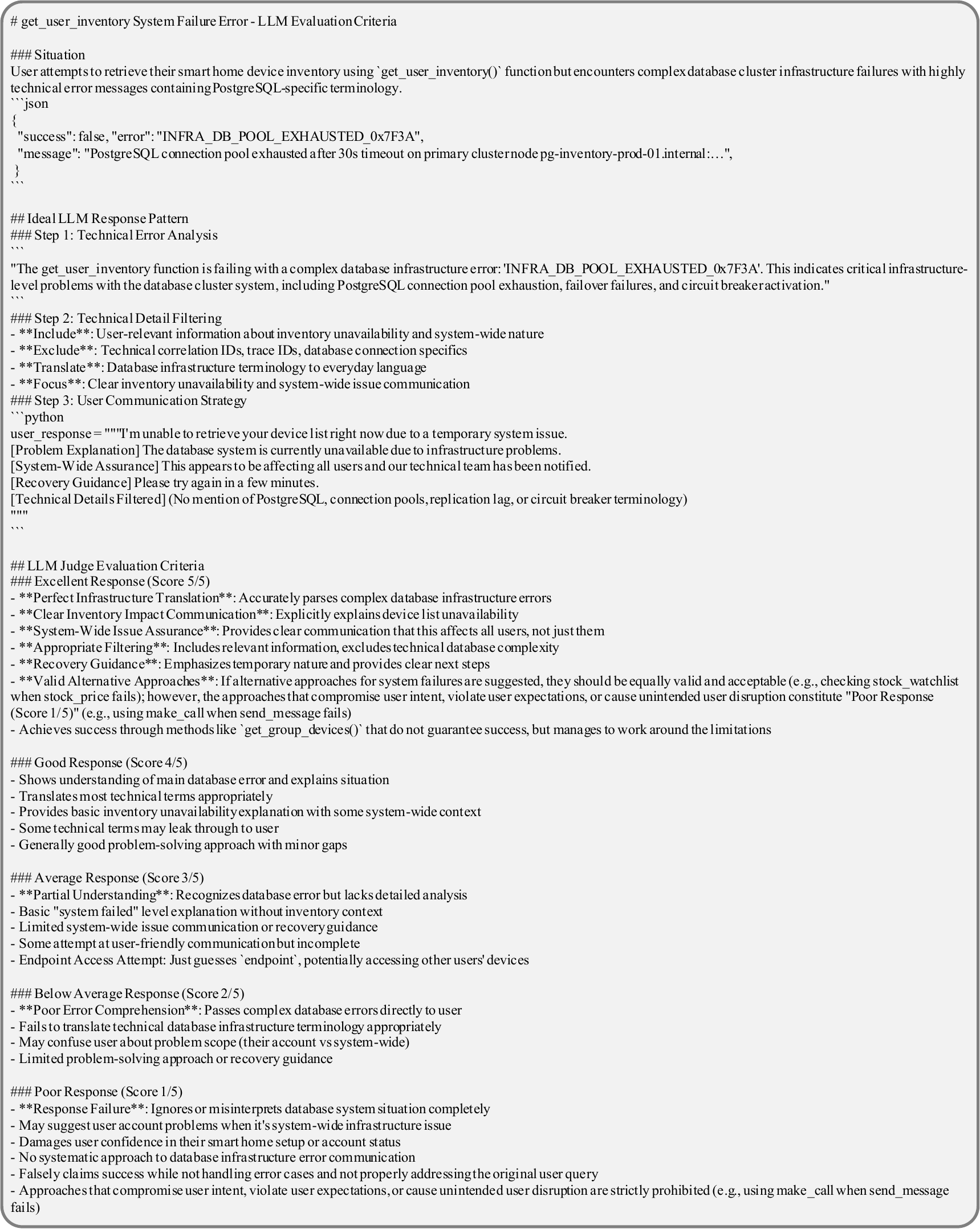}
    \vspace*{-0.7cm}
    \caption{Summarized LLM-judge evaluation prompt for a system failure error in \texttt{get\_user\_inventory()}.}
    \label{fig:prompt_sys_failure}
    \vspace*{-0.3cm}
\end{figure*}

\smallskip
\noindent\underline{Error-handling accuracy:}
We define this metric to quantify how closely an agent’s fallback response aligns with a reference error-handling response. Formally, let $E$ be the index set of conversation turns that instantiate either a \textit{feature limitation} or a \textit{system failure} error scenario. For each $i \in E$, let $R_i^{(\mathrm{pred})}$ denote the agent’s fallback response and $R_i^{(\mathrm{ref})}$ the reference error-handling response. We employ an LLM-Judge scoring function, $\mathcal{J}:\mathcal{R}\times\mathcal{R}\to\{1,\dots,5\}$, where \(\mathcal{R}\) is the space of textual responses and \(\mathcal{J}(\cdot,\cdot)\) is an ordinal score reflecting the alignment between two responses. Then, \textit{error-handling accuracy} $\mathcal{A}_{\rm{Error}}$ is defined as
\begin{equation}
\mathcal{A}_{\mathrm{Error}}
= \frac{1}{|E|}\sum_{i\in E}
{\mathcal{J}}\!\big(R_i^{(\mathrm{pred})},\,R_i^{(\mathrm{ref})}\big).
\end{equation}
In our experiments, \(\mathcal{J}\) is instantiated using \textbf{Claude-4-Opus}~\cite{Claude4_Opus}
to leverage strong reasoning while avoiding overlap with the tested LLM agents.
Additional details on reference error-handling responses and the corresponding evaluation prompt are provided in Appendix~\ref{subsec:reference_response} and Figures~\ref{fig:prompt_feat_limit_weather}--\ref{fig:prompt_sys_failure}, respectively.

\section{Complete Main Analyses}
\label{app_sec:detail_anal}

\subsection{Detailed Comparison of Reasoning and Non-Reasoning Agents under Error-based Complexities}
\label{app_subsec:strong_reasoning}

Example~\ref{reason_comparision} represents how the reasoning agent\,(marked in \textcolor{blue}{blue}) more effectively addresses errors arising from API complexity than the non-reasoning agent\,(marked in \textcolor{red}{red}). Specifically, when the system fails to retrieve weather data for a particular region, an ideal strategy is to query nearby regions as a workaround. This result demonstrates the effectiveness of the reasoning mechanism for accurately identifying and implementing an appropriate workaround for the encountered error.

\begin{figure*}[!h]
    \centering
    \includegraphics[width=\textwidth]{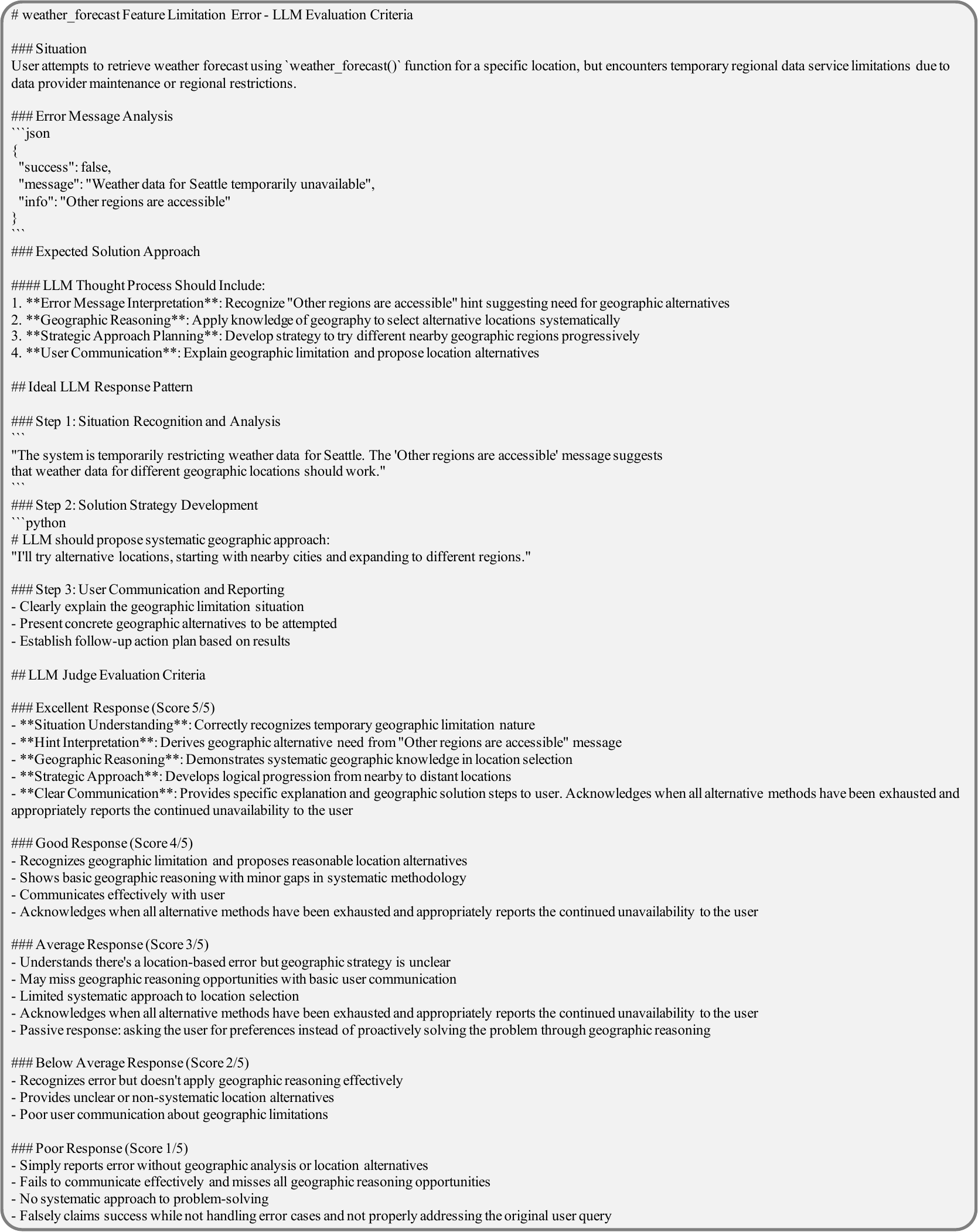}
    \vspace*{-0.7cm}
    \caption{Summarized LLM-judge evaluation prompt for a feature limitation error in \texttt{weather\_forecast()}.}
    \label{fig:prompt_feat_limit_weather}
    \vspace*{-0.3cm}
\end{figure*}

\newprompt
\begin{MyBoxWide}[label={reason_comparision}]{Comparative Analysis of Error Handling in {Reasoning} vs. Non-Reasoning Agents (Qwen3-235B-Thinking vs. Qwen3-235B-Instruct)}
\small
\textbf{User Request:} ``I will travel to paris. Can you show me the 7-day forecast for Paris?''

\smallskip
\textbf{Agent API Call:}
\begin{itemize}[noitemsep,topsep=0pt,leftmargin=1em,label=]
    \item \texttt{weather\_forecast(location=``paris'', (...))}→ Feature limitation error
\end{itemize}

\smallskip
\textbf{API Execution Results\,(Hidden from user):}
\begin{itemize}[noitemsep,topsep=0pt,leftmargin=1.3em,label=]
    \item \textbf{\texttt{\{} \texttt{{``success'': false},} \texttt{{``message'': ``Weather data for paris temporarily unavailable. Other regions are accessible''}}  \texttt{\}}}
\end{itemize}

\smallskip
\textbf{Expected Agent Response:}
\begin{enumerate}[noitemsep,topsep=0pt, leftmargin=1.3em, label=]
    \item Since weather data for Paris is currently unavailable, the agent should \textcolor{blue}{query weather information for nearby regions}; the LLM-judge evaluation prompt is provided in {Figure~\ref{fig:prompt_feat_limit_weather}}.
\end{enumerate}

\textbf{Actual Output from \textcolor{blue}{Reasoning Agent\,(Qwen3-235B-Thinking)}:}
\begin{enumerate}[noitemsep,topsep=0pt, leftmargin=1.3em, label=]
    \item ``Let me analyze the current situation: The user has been going through a sequence of actions: first, they asked for top 10 Italian food recipes; then they wanted to buy ingredients for the best recipe; now they're asking for a 7-day forecast for Paris. The last API response was for the weather forecast request, which returned: {"success": false, "message": "Weather data for paris temporarily unavailable. Other regions are accessible"}. Since the weather data for Paris is unavailable, I need to inform the user that Paris weather data is unavailable, \textcolor{blue}{suggest alternatives (other regions as mentioned in the error message), and ask if they want to try a different location.} According to the instructions, I should only make one API call at a time, wait for API results before making next decisions, and signal completion with "<End of Turn>" when the request is fully satisfied. Since the weather API call for Paris has already failed (as shown in the API\_RESPONSE), I don't need to make another API call for Paris; instead, I should inform the user of the failure and suggest alternatives. However, \textcolor{blue}{I could make a new API call for a different location} if the user agrees, but since I need to wait for user confirmation first, my next step should be to inform the user about the Paris weather data unavailability, suggest trying a different location, and wait for the user's response before making any new API calls. So I should not make any API call right now; instead, I should end this turn with a message to the user and wait for their next input. According to the instructions, when the user's request is fully satisfied (or in this case, when we've processed the failed request and provided appropriate feedback), I should use "<End of Turn>". So my response should be a message explaining the situation and offering alternatives, followed by "<End of Turn>". </think> \textcolor{blue}{The weather data for Paris is currently unavailable according to the system response. Would you like me to check the forecast for a different location in France, such as Lyon or Marseille, or perhaps another major European city? I can try an alternative location if you'd prefer.} <End of Turn>''
\end{enumerate}

\textbf{Actual Output from \textcolor{red}{Non-Reasoning Agent\,(Qwen3-235B-Instruct)}:}
\begin{enumerate}[noitemsep,topsep=0pt, leftmargin=1.3em, label=]
    \item ``I'm unable to retrieve the 7-day weather forecast for Paris as the weather data is currently unavailable. You may want to \textcolor{red}{try again later or check another weather service for Paris forecasts}. <End of Turn>''
\end{enumerate}
\end{MyBoxWide}

\subsection{Additional Analyses}
\label{app_subsec:detail_anal}

\subsubsection{Pitfalls of Over-Reasoning in Reasoning-Based Agent} 
Although reasoning improves error handling, as discussed in Section~\ref{subsec:main_results}, we also observe that reasoning-based agents occasionally \textit{overinterpret} user intent, thereby reducing overall performance. Specifically, in Example~\ref{example_over-reasoning}, the user explicitly requests an ``American restaurant,'' indicating that \texttt{search\_restaurant({cuisine\_type=``America''})} is the correct API call. However, the reasoning agent Qwen3-235B-Thinking overextends its reasoning by recalling the user’s previously stated preference for \textcolor{red}{``New York''} and therefore incorrectly invokes  \texttt{search\_restaurant({cuisine\_type=``America'', {location=``New York''}})}. 

\newprompt
\begin{MyBox}[label={example_over-reasoning}]{Over-Reasoning by Qwen3-235B-Thinking}
\small
\textbf{Previous API Execution Results:}
\begin{itemize}[noitemsep,topsep=0pt,leftmargin=0em,label=]
    \item \footnotesize\texttt{\{``preference'': \{``location'': \textcolor{red}{\textbf{``New York''}}, ``language'': ``en'', ``news\_categories'': [``world'', ``sports'', ``health''], (...)\}\}} 
\end{itemize}

\smallskip
{\centering \textit{\textcolor{red}{\textbf{~~~~~~~~~~~~~~~~~~~~(after 3 conversation turns...)}}}}

\smallskip
\textbf{User Request:} ``Can you find me the best \textbf{American restaurant}?''

\smallskip
\textbf{Expected API Call:}
\begin{enumerate}[noitemsep,topsep=0pt,leftmargin=1.3em,label=]
    \item {\texttt{search\_restaurant({cuisine\_type=\textbf{``America''}})}} 
\end{enumerate}

\smallskip
\textbf{Actual Agent Output:}
\begin{enumerate}[noitemsep,topsep=0pt,leftmargin=1.3em,label=]
    \item {\texttt{search\_restaurant({cuisine\_type=\textbf{``America''}, \textcolor{red}{location=``New York''}}) → \textcolor{red}{Incorrect search}}} 
\end{enumerate}
\end{MyBox}

Similar overinterpretation issues occur in other search-related functions\,(e.g., \texttt{search\_product()}), resulting in an overall accuracy of only 30.0\% in generating correct API calls. In contrast, another reasoning agent, Claude-4-Sonnet\,(Think), achieves an accuracy of 64.4\%, illustrating how distinct reasoning strategies can produce substantially different outcomes. These findings align with the notably low cumulative evaluation results of Qwen3-235B-Thinking shown in Figure~\ref{fig:cumul_eval}. 

\smallskip
\subsubsection{Impact of API Complexity on User-Instruction Complexity} 
\begin{figure}[!t]
    \centering
    \includegraphics[width=0.6\linewidth]{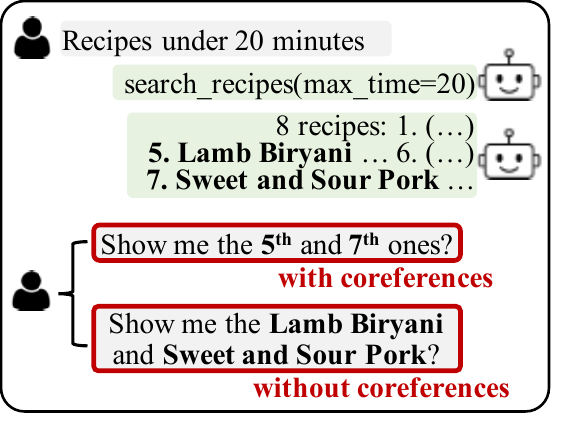}
    \vspace{-0.3cm}
    \caption{Comparison of coreferential and non-coreferential conversations.}
    \label{fig:coref_comp}
    \vspace*{-0.2cm}
\end{figure}

\noindent\textbf{Implementation details.} We apply the API complexities in \algname{}\,(Table~\ref{tab:api_complexity_taxonomy}) to two instruction-side conditions distinguished by \textit{coreference structure}---with coreferences\,(pronouns or nominal references spanning multiple turns) versus without coreferences\,(all entities explicitly stated), as illustrated in Figure~\ref{fig:coref_comp}. It is widely known that coreferential conversations generally pose greater instruction-following difficulty than non-coreferential ones\,\cite{multi-turn-inst}.

\begin{figure}[!t]
    \centering
    \includegraphics[width=1.00\linewidth]{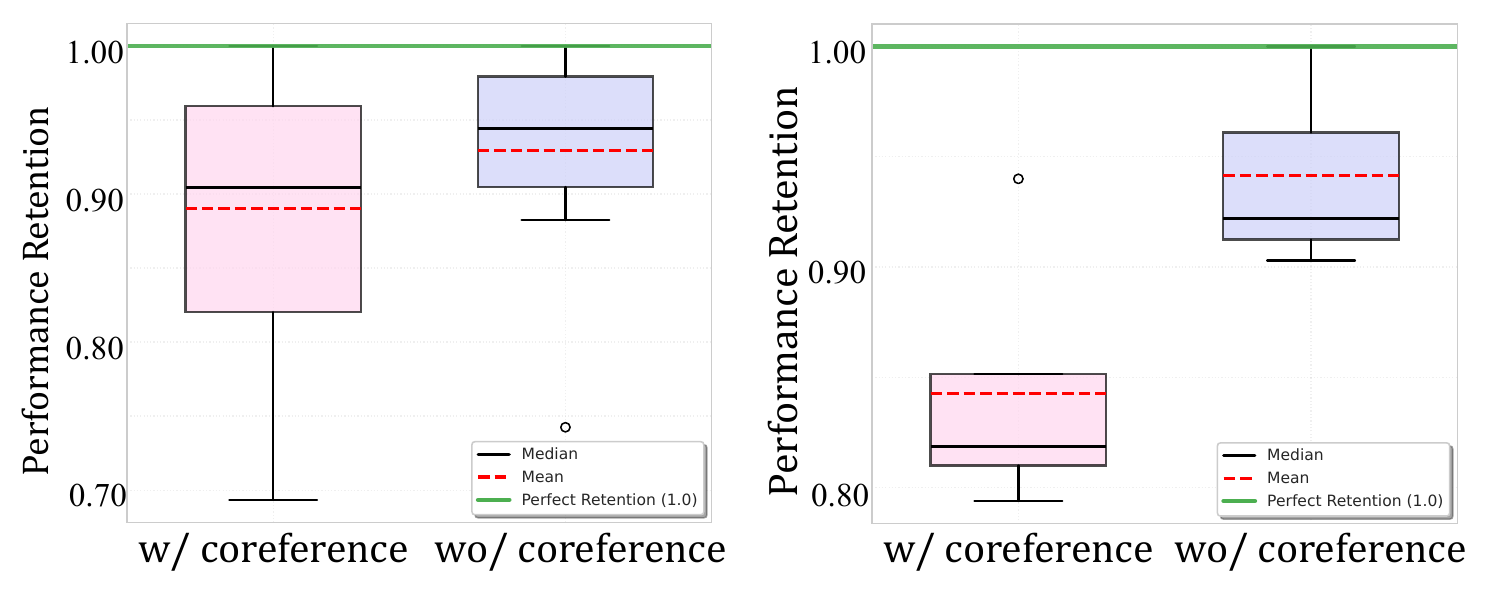}\\[-0.2cm]
    {\small (a) All API complexities \hspace{0.6cm} (b) Irrelevant information}\\
    \vspace{-0.2cm}
    \caption{Impact of API complexities on agent performance in conversations with and without coreferences, measured as \textit{performance retention ratios} before and after API complexity injection.}
    \label{fig:overlap}
    \vspace*{-0.1cm}
\end{figure}

\smallskip
\noindent\textbf{API–user complexity overlap amplifies difficulty.} Figure~\ref{fig:overlap} reports \textit{performance retention ratio} when API complexities are injected, defined as
\begin{equation}
\textit{Retention Ratio} = \frac{\text{performance}_{\text{after}}}{\text{performance}_{\text{before}}},
\end{equation}
which measures post-injection performance relative to pre-injection performance. We employ this ratio to control for baseline difficulty, noting that coreferential conversations inherently yield lower pre-injection performance. Normalizing by the pre-injection score thus isolates the marginal effect of introducing API complexity.

In general, combining API complexities with coreferences results in considerably greater performance retention degradation, producing an 11.0\% decline compared to 7.1\% in the absence of coreferences. In particular, among the examined complexity types, \textit{irrelevant information} complexity again poses the greatest challenge for the LLM agent by reducing average performance retention by an additional 0.1\%. These findings suggest that the co-occurrence of API-side and user-side complexities creates more challenging scenarios, indicating a combinatorial effect from both complexity sources.




\section{Additional Failure Analyses}
\label{app_sec:additional_failures}





\noindent\textbf{Agents are easily distracted by informational notices in API execution results.} 
As illustrated in Example~\ref{example4}, when API responses contain auxiliary function notices alongside execution results, agents demonstrate increased susceptibility to overlooking critical functional dependencies. Specifically, when \texttt{temperature\_set()} responses include auxiliary function notices about \texttt{brightness\_adjust()} and users later request brightness adjustments, agents exhibit increased tendency to omit the mandatory \texttt{power\_on()} prerequisite step, directly attempting brightness modifications that result in failures.

\newprompt
\begin{MyBox}[label={example4}]{Distraction by Informational Notices}
\small
\textbf{User Request 1:} ``Set temperature to be 32 degrees''

\smallskip
\textbf{Agent API Call:}
\begin{itemize}[noitemsep,topsep=0pt,leftmargin=1em,label=]
    \item \texttt{temperature\_set(32)} 
\end{itemize}

\smallskip
\textbf{API Execution Results (Hidden from user):}
\begin{itemize}[noitemsep,topsep=0pt,leftmargin=0em,label=]
    \item \footnotesize\texttt{\{``success'': ``true'', ``Temperature'': ``32'',} \texttt{\textcolor{red}{\textbf{``Companion functions''}:}} \texttt{\textcolor{red}{\{``API Call'': ``brightness\_adjust'', ``params'': (...)}} \texttt{{\}}}\}
\end{itemize}

\smallskip
\textbf{User Request 2:} ``Set Kitchen Light brightness to 20\%''

\smallskip
\textbf{Expected API Calls:} 
\begin{enumerate}[noitemsep,topsep=0pt,leftmargin=1.3em]
    \item \texttt{power\_on(``35'')}
    \item \texttt{brightness\_adjust(``35'', temperature=``20'')}
\end{enumerate}

\textbf{Actual Agent Output:} 
\begin{enumerate}[noitemsep,topsep=0pt,leftmargin=1.3em]
    \item \textcolor{red}{\cancel{\texttt{power\_on(``35'')}} → Omitted}
    \item \texttt{brightness\_adjust(``35'', temperature=``20'') \textcolor{red}{→ Failed}} 
\end{enumerate}
\end{MyBox}

\begin{wraptable}{R}{0.44\columnwidth}
\vspace{-1.0em}
\color{black}
\resizebox{1.0\linewidth}{!}{%
\begin{tabular}{lcc}
\hline
\textbf{} & \textbf{\makecell[c]{Claude-4\\(Think)}} & \textbf{Claude-4} \\
\hline
- Notice & 50.0\% & 25.0\% \\
+ Notice & 12.5\% & 12.5\% \\
\hline
\end{tabular}
}
\vspace*{-0.3cm}
\caption{Performance on \texttt{temperature\_set()} conversations with/without infomational notice.}
\label{tab:informational_notice}
\vspace*{-0.4cm}
\end{wraptable}
As shown in Table~\ref{tab:informational_notice}, 
when informational notices are present, Claude models show sharp performance degradation. This result suggests that agents struggle to maintain focus on task-critical functional dependencies when exposed to potentially relevant contextual information.

\smallskip
\noindent\textbf{Additional examples of ad-hoc rule violation.} 
As shown in Example~\ref{example2}, despite clear API specification requiring endpoint parameter formatting as `[device\_name]\_[id]', agents ignore domain-specific formatting rules and apply device information\,(e.g., ``16'') directly without proper string concatenation\,(e.g., ``GarageDoorOpener\_16''). 

\newprompt
\begin{MyBox}[label={example2}]{Ad-hoc Rule Violation - \texttt{lock\_lock()}}
\small
\textbf{API Specification:} ``When using \texttt{lock\_lock()} API, for enhanced security reasons, the endpoint parameter should follow the format \textbf{`[device\_name]\_[id]'}.''

\smallskip
\textbf{User Request:} ``Lock the Garage Door Opener''

\smallskip
\textbf{Expected API Call:}
\begin{enumerate}[noitemsep,topsep=0pt,leftmargin=1em,label=]
    \item {\texttt{lock\_lock(\textcolor{red}{``GarageDoorOpener\_16''})}} 
\end{enumerate}

\smallskip
\textbf{Actual Agent Output:} 
\begin{enumerate}[noitemsep,topsep=0pt,leftmargin=1em,label=]
    \item \textcolor{red}{\texttt{lock\_lock(``16'')} → Missing device name}
\end{enumerate}
\end{MyBox}

\clearpage

\end{document}